\documentclass[sigconf]{acmart}

\usepackage{multirow}
\usepackage{url}
\usepackage{subfigure}
\usepackage{float}
\usepackage{makecell}

\AtBeginDocument{%
  \providecommand\BibTeX{{%
    \normalfont B\kern-0.5em{\scshape i\kern-0.25em b}\kern-0.8em\TeX}}}

\begin{document}

\title{Which Kind Is Better in Open-domain Multi-turn Dialog, Hierarchical or Non-hierarchical Models? An Empirical Study}


\author{Tian Lan}
\email{lantiangmftby@gmail.com}
\affiliation{%
  \institution{Beijing Institute of Technology}
  \city{Beijing}
  \state{Beijing}
  \postcode{100081}
}

\author{Xian-ling Mao}
\email{maoxl@bit.edu.cn}
\affiliation{%
  \institution{Beijing Institute of Technology}
  \city{Beijing}
  \state{Beijing}
  \postcode{100081}
}

\author{Wei Wei}
\email{Weiw@hust.edu.cn}
\affiliation{%
  \institution{Huazhong University of Science and Technology}
  \city{Wuhan}
  \state{Hubei}
  \postcode{430074}
}

\author{Heyan Huang}
\email{hhy63@bit.edu.cn}
\affiliation{%
  \institution{Beijing Institute of Technology}
  \city{Beijing}
  \state{Beijing}
  \postcode{100081}
}

\begin{abstract}

  Currently, open-domain generative dialog systems have attracted considerable attention in academia and industry. 
  Despite the success of single-turn dialog generation, multi-turn dialog generation is still a big challenge. 
  So far, there are two kinds of models for open-domain multi-turn dialog generation: hierarchical and non-hierarchical models. 
  Recently, some works have shown that the hierarchical models are better than non-hierarchical models under their experimental settings;
  meanwhile, some works also demonstrate the opposite conclusion. 
  Due to the lack of adequate comparisons, 
  it's not clear which kind of models are better in open-domain multi-turn dialog generation. 
  Thus, in this paper, we will measure systematically nearly all representative hierarchical and non-hierarchical models over the same experimental settings to check which kind is better. 
  Through extensive experiments, we have the following three important conclusions: 
  (1) Nearly all hierarchical models are worse than non-hierarchical models in open-domain multi-turn dialog generation, except for the HRAN model. Through further analysis, the excellent performance of HRAN mainly depends on its word-level attention mechanism;
  (2) The performance of other hierarchical models will also obtain a great improvement if integrating the word-level attention mechanism into these models. The modified hierarchical models even significantly outperform the non-hierarchical models;
  (3) The reason why the word-level attention mechanism is so powerful for hierarchical models is because it can leverage context information more effectively, especially the fine-grained information. 
  Besides, we have implemented all of the models and already released the codes\footnote{The anonymous link: \url{https://github.com/anonymous/xxx}}.

\end{abstract}


\keywords{Multi-turn; Open-domain dialog generation; Hierarchical models}


\maketitle

\section{Introduction}
Recently, open-domain generative dialog systems are attracting
increasing attention due to its promising potentials and alluring commercial values.
With the huge development of deep learning, neural networks can generate
a very fluent response based on one user's utterance, which is usually called the open-domain single-turn dialog generation, 
such as CCM \cite{Zhou2018CommonsenseKA} and DialoGPT \cite{Zhang2019DialoGPTLG}.
However, the daily conversations between two humans are actually the multi-turn manner, 
which contains multiple utterances as the conversation context (dialog history). 
Unlike the single-turn manner, the multi-turn dialog models need to make full use of the multiple utterances for generating a coherent response.
For example, as shown in Table \ref{tab:1}, the single-turn dialog models can only focus the last utterance in the conversation context, 
which leads to the bad responses in the second conversation. 
The suitable and coherent responses can only be generated by fully considering the information of the multiple utterances,
which is still a big challenge.

\begin{table}[H]
  \begin{center}
  \begin{tabular}{|c|c|c|}
  \toprule[2pt]
  \hline
  \multirow{3}{*}{\textbf{(1)}} & \textbf{Context}     & \begin{tabular}[c]{@{}c@{}}Hey, I got a perfect score on this test.\\ Wow, wonderful !\\ How should i tell my mom ?\end{tabular} \\ \cline{2-3} 
                                & \textbf{Single-turn} & Give her a surprise !                                                                                                            \\ \cline{2-3} 
                                & \textbf{Multi-turn}  & Give her a surprise !                                                                                                            \\ \hline \hline
  \multirow{3}{*}{\textbf{(2)}} & \textbf{Context}     & \begin{tabular}[c]{@{}c@{}}Oh, I failed the exam.\\ That sounds terrible !\\ How should i tell my mom ?\end{tabular} \\ \cline{2-3} 
                                & \textbf{Single-turn} & Give her a surprise !                                                                                                            \\ \cline{2-3} 
                                & \textbf{Multi-turn}  & Tell her the truth. Next time, keep it up !                                                                                      \\ \hline
  \bottomrule[2pt]
  \end{tabular}
  \caption{Cases of the multi-turn and single-turn manners.}
  \label{tab:1}
  \end{center}
\end{table}

\begin{figure}[t]
  \centering
  \subfigure[Hierarchical architecture.]{
      \begin{minipage}[t]{\linewidth}
          \centering
          \includegraphics[width=7cm, height=5cm]{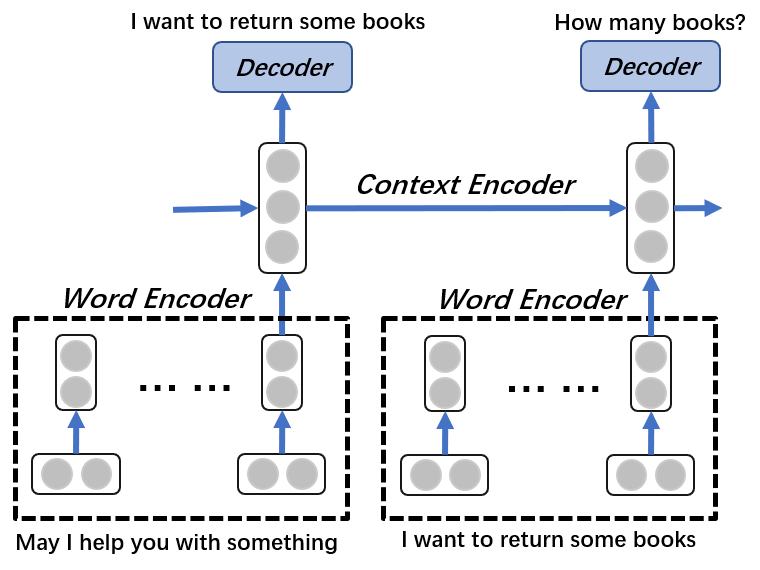}
      \end{minipage}%
      \label{img:1a}
  }%
  
  \subfigure[Non-hierarchical architecture.]{
      \begin{minipage}[t]{\linewidth}
          \centering
          \includegraphics[width=7cm, height=3.5cm]{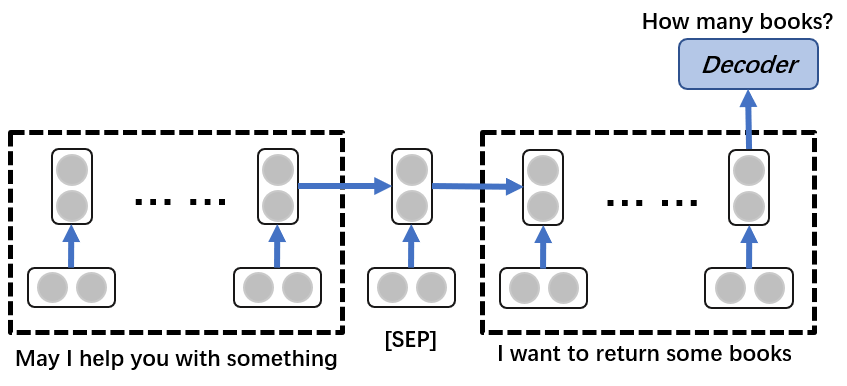}
      \end{minipage}%
      \label{img:1b}
  }%
  \centering
  \caption{Hierarchical architecture and non-hierarchical architecture for the multi-turn dialogue modeling. The word encoder, context encoder and decoder are usually the RNN-based neural networks.
  The \textbf{[SEP]} token means the separator between two utterances.}
  \label{img:1}
\end{figure}

So far, there are two kinds of models for modeling multi-turn conversations: hierarchical and non-hierarchical models.
For hierarchical models, as shown in Figure 1 (a), they usually contain two encoders and one decoder: 
(1) Word encoder expresses one sentence in a multi-turn conversation as a dense vector, which represents the semantics of the input message. The GRUs \cite{Bahdanau2014NeuralMT} and LSTMs \cite{Hochreiter1997LongSM} are most commonly used \cite{Zhang2019ReCoSaDT}.
(2) Context encoder captures the context-level information based on the semantics of the utterances. The RNNs \cite{Bahdanau2014NeuralMT} and Transformers architecture \cite{Vaswani2017AttentionIA} are most commonly used in recent works \cite{Zhang2018ContextSensitiveGO,Zhang2019ReCoSaDT}.
(3) Decoder module finally generates a context-sensitive response according to the semantic representation of the multi-turn conversation.
For non-hierarchical models, as shown in Figure 1 (b), they leverage the basic Seq2Seq encoder-decoder architecture \cite{Sutskever2014SequenceTS} to 
directly maps the input sequence to the target sequence by using deep neural networks.
In multi-turn settings, the researchers usually simply concatenate the multiple sentences with a separator.

Recently, some works have shown that the hierarchical models have a more powerful capability to leverage the multi-turn context than non-hierarchical models \cite{Tian2017HowTM,Xing2017HierarchicalRA,Zhang2019ReCoSaDT}.
Meanwhile, the experiments of some works \cite{Wang2018ChatMD,Xu2019DynamicWM} demonstrate the opposite conclusion, i.e. the hierarchical models are worse than non-hierarchical models under their experimental settings.
Due to the lack of systematic comparisons, 
it is still not clear which kind of models are better in open-domain multi-turn dialog generation.
Thus, in this paper, we will measure systematically nearly all representative hierarchical and non-hierarchical models over the same experimental settings to check which kind is better.


From extensive experiments, we obtain three important conclusions:
(1) Nearly all of the hierarchical models are worse than the non-hierarchical models in open-domain multi-turn dialog generation, except for the HRAN model \cite{Xing2017HierarchicalRA}.
Through further analysis, the excellent performance of HRAN model mainly depends on its word-level attention mechanisms, 
which is usually ignored by the recent researches \cite{Zhang2019ReCoSaDT,Zhang2018ContextSensitiveGO,Chen2018HierarchicalVM};
(2) The performance of other existing hierarchical models will also obtain a great improvement if integrating the word-level attention mechanism into them.
Besides, the modified hierarchical models even significantly outperform the non-hierarchical models;
(3) The reason why the word-level attention mechanism is so powerful for hierarchical models is because
it can leverage the context information more effectively, especially the fine-grained information.

In this paper, our main contributions are three-fold:
\begin{itemize}
  \item It is the first time to study the fundamental question of which kind of models are better in open-domain multi-turn dialog generation.
  We systematically compare the existing hierarchical and non-hierarchical models in open-domain multi-turn dialog generation.
  \item Extensive experiments demonstrate three important conclusions: 
  (1) \textbf{Nearly all hierarchical models are worse than the fundamental non-hierarchical models, except for HRAN model};
  (2) \textbf{Word-level attention mechanism greatly improves the performance of hierarchical models}, which is usually ignored by the existing researches.
  The modified hierarchical models even significantly outperform the state-of-the-art non-hierarchical models;
  (3) Quantitative and qualitative analysis demonstrate that \textbf{the word-level attention mechanism can help the hierarchical models leverage the multi-turn context more effectively, especially the fine-grained information}.
  \item \textbf{We have implemented all of the chosen models and already released the codes}, which will be quite helpful for the dialog system research community.
\end{itemize}


\section{Related Work}
\subsection{Generative Multi-turn Dialog} 
Despite the success of single-turn dialog generation, multi-turn dialog generation is still a big challenge.
So far, there are two kinds of models for open-domain multi-turn dialog generation: hierarchical and non-hierarchical models.

\noindent \textbf{Hierarchical Models:}
The most important work of hierarchical models is the HRED model \cite{Serban2015BuildingED}, which contains the word-level and utterance-level encoders:
(1) Word-level encoder mainly focuses on representing utterances by using RNN \cite{Cho2014LearningPR};
(2) Utterance-level encoder leverages the utterance representation generated by the word-level encoder to capture the session-level information.
Based on HRED, lots of hierarchical models are proposed, and the main architectures of these proposed models are consistent with the HRED.
First, WSeq \cite{Tian2017HowTM} is a hierarchical model that uses the cosine similarity to weight the utterance representations generated by utterance-level encoder.
Then, VHRED \cite{Serban2016AHL} introduces the latent variable into HRED model for generating more diverse responses.
Furthermore, HRAN model \cite{Xing2017HierarchicalRA} contains the hierarchical attention mechanism (word-level and utterance-level attention) for HRED.
Moreover, DSHRED model \cite{Zhang2018ContextSensitiveGO} is proposed for generating context-sensitive responses for HRED, 
which uses the dynamic and static attention mechanisms to focus the last utterance in the conversation.
Recently, ReCoSa model \cite{Zhang2019ReCoSaDT} replaces the RNN-based utterance-level encoder with the multi-head self-attention module to detect multiple relative sentences in the conversation context,
and shows the state-of-the-art performance.

\noindent \textbf{Non-hierarchical Models:} 
The motivation of the non-hierarchical models is to simplify the multi-turn dialog modeling by the single-turn manner, i.e., simply concatenate multiple sentences into one sentence.
The non-hierarchical models usually take use of the basic Seq2Seq with attention architecture.
The NRM model \cite{Shang2015NeuralRM} is the first non-hierarchical model for dialog generation, which has the RNN-based encoder and decoder modules. 
Recently, the transformer architecture \cite{Vaswani2017AttentionIA} shows more powerful capability than RNN models for modeling the long sequence,
which is very suitable for processing the multi-turn context,
and some works have been proposed to use transformer model for open-domain dialog generation, such as DialoGPT \cite{Zhang2019DialoGPTLG} and Meena \cite{Adiwardana2020TowardsAH}.

Some researches show that the hierarchical models are better than the non-hierarchical models for multi-turn dialogue modeling \cite{Serban2015BuildingED,Serban2016AHL,Tian2017HowTM,Xing2017HierarchicalRA,Zhang2019ReCoSaDT,Zhang2018ContextSensitiveGO}.
Meanwhile, there are also some works obtaining the opposite conclusion under their experimental settings \cite{Wang2018ChatMD,Xu2019DynamicWM}.
Due to the lack of systematic comparisons between hierarchical and non-hierarchical models, 
it is still not clear which kind of models are better.
So we conduct extensive experiments to research the problem in this paper.

\subsection{Attention Mechanism}
In natural language generation, the attention mechanism is usually used to improve the performance of the encoder-decoder architecture,
which can provide high-quality context representation (context vector) for decoding.
The attention mechanism can be described as mapping a query and a set of key-value pairs to a context vector $c$ \cite{Vaswani2017AttentionIA}, where the query, keys, values, and $c$ are all vectors. 
$c$ is computed as a weighted sum of the values, where the weight assigned to each value is computed by an attention module of the query with the corresponding key.

For non-hierarchical models, the context vector $c$ is calculated as follows:
\begin{equation} \label{form:1}
  e_{ij} = \rm{attn}(s_i,h_j), w_{ij}=\frac{\rm{exp}(e_{ij})}{\sum_k exp(e_{ik})}
\end{equation} where the $s_i$ is the hidden state of the decoder model in step $i$, which is also the query.
$h_j$ is the hidden state of the encoder model, which is also the $j$-th key (there are $k$ key-value pairs).
In most of the works, the $\rm{attn}$ module is a one-layer neural network, which generates the weight score $e_{ij}$.
Then, the \textit{softmax} function is used to normalize all weights. 
Finally, the context vector $c$ can be represented:
\begin{equation} \label{form:2}
  c_i = \sum_k w_{ik}h_k
\end{equation} where the $h_k$ is the value (same as the key).

For hierarchical models, there are two kinds of attention mechanisms \cite{Xing2017HierarchicalRA}: word-level attention and utterance-level attention.
It should be noted that the calculation process of the two kinds of attention mechanisms are the same as the one in Formula (1) and Formula (2).
The context vector $c_i$ of hierarchical models is obtained by the following steps:
First of all, suppose the multi-turn context contains $m$ utterances, $m$ different context vectors will be obtained $\{c_{ij}\}_{j=1}^m$.
Then, these context vectors $\{c_{ij}\}_{j=1}^m$ will be fed as the input vector to the context encoder model, 
and the hidden state of the context encoder $\{H_{ij}\}_{j=1}^m$ are obtained.
Finally, the utterance-level attention is used to obtain the final context vector $c_i$ for decoding based on $\{H_{ij}\}_{j=1}^m$.

After obtaining the context vector $c_i$, the decoder will generated the context-sensitve responses,
and the $(i+1)$-th token $t_{i+1}$ can be generated:
\begin{equation}
  t_{i+1} = f(s_i;c_i;t_i)
\end{equation} where the function $f$ is the RNNs or transformer model.

\section{Systematic comparisons}
In this section, we will show the details of our systematic comparisons of the existing models for open-domain multi-turn dialog generation.
First of all, the experimental settings are shown in Section 3.1.
Then, we systematically compare the representative hierarchical and non-hierarchical models on four open-domain dialog datasets in Section 3.2.
Moreover, in order to check whether the word-level attention mechanism has the consistent improvement for hierarchical models,
it is added into other hierarchical models, and the results are shown in Section 3.3.
Finally, in Section 3.4 and 3.5, the quantitative and qualitative analysis are elaborated to analyze why word-level attention mechanism is so effective.

Due to the page limitation, we only show the partial results in this paper, 
and more details can be found in the GitHub repository\footnote{Anonymous link, \url{https://github.com/g32M7fT6b8Y/External-Experiments}}. 
Noted that the conclusions over these partial results in this paper are consistent with all the results.

\subsection{Experimental Setting}
\subsubsection{Chosen Datasets}
In order to systematically compare the models, we choose four popular English open-domain multi-turn dialog datasets:
\begin{itemize}
  \item DailyDialog \cite{Li2017DailyDialogAM}: DailyDialog is a high-quality open-domain multi-turn dialogue corpus which covers various topics about our daily life.
  Recently, lots of researches use this dataset to evaluate their models \cite{Wang2018ChatMD,rashkin2019towards}.
  \item EmpChat \cite{rashkin2019towards}: EmpChat is a new benchmark for empathetic dialog generation, called EmpatheticDialogues, a novel dataset of 25k conversations grounded in emotional situations.
  In this work, we simply ignore the labels and situations information.
  \item DSTC7-AVSD \cite{AlAmri2019AudioVS}: DSTC7-AVSD dataset contains multimodal conversations such as video and text. 
  This task aims to generate a complete and natural response to a question about a scene, given video and audio of the scene and the history of previous turns in the dialog. 
  Here, we just simply ignore the video and audio modal and construct the corpus by only using the text.
  \item PersonaChat \cite{Zhang2018PersonalizingDA}: PersonaChat is the first work to introduce profile information (sentences about the persona) as the condition to maintain the consistent personality during dialog generation.
  In this work, the persona sentences of the speaker are placed at the forefront of the conversation context.
\end{itemize}
These datasets are carefully pre-processed, and the statistics about processed datasets can be found in Table \ref{tab:2}.

\begin{table}[h]
  \begin{center}
    \resizebox{0.48\textwidth}{!}{
      \begin{tabular}{|c|c|c|c|c|c|c|c|}
      \hline
      \multirow{2}{*}{\textbf{Dataset}} & \multicolumn{3}{c|}{\textbf{Turn}}         & \multicolumn{3}{c|}{\textbf{Length}}       & \multirow{2}{*}{\textbf{Vocab}} \\ \cline{2-7}
                                        & \textbf{max} & \textbf{avg} & \textbf{min} & \textbf{max} & \textbf{avg} & \textbf{min} &                                 \\ \hline
      \textbf{DailyDialog}              &    34        &   5.09       &   1          &     262      &   14.46      &     2        &      23,869                           \\ \hline
      \textbf{PersonaChat}              &    54        &  11.96       &   4          &     61       &   11.02      &     2        &      18,096                           \\ \hline
      \textbf{DSTC7-AVSD}               &    36        &  13.26       &   2          &     69       &   11.27      &     2        &      10,850                          \\ \hline
      \textbf{EmpChat}                  &     7        &   2.23       &   1          &     111      &   14.95      &     2        &      37,109                           \\ \hline
      \end{tabular}
    }
    \caption{The statistics of the chosen datasets. The number of turns in the multi-turn conversation, 
    and the length of each utterance are reported.}
    \label{tab:2}
  \end{center}
\end{table}

\subsubsection{Chosen Models}
We implement the nearly all representative hierarchical dialogue models and non-hierarchical models in consistent settings for systematic comparisons.

\noindent \textbf{Hierarchical Models}: 
\begin{itemize}
  \item HRED \cite{Serban2015BuildingED}: HRED is proposed in \citeyear{Serban2015BuildingED}, which is the first work to use the hierarchical encoder-decoder architecture to model the multi-turn dialog generation. 
  Here we enhance the vanilla HRED by adding the utterance-level attention module to the context encoder.
  \item VHRED \cite{Serban2016AHL}: In \citeyear{Serban2016AHL}, VHRED is proposed to add the latent stochastic variables before decoding for the HRED context encoder,
  which aims to generate more diverse utterances. In our implementation, the KL annealing method \cite{Bowman2015GeneratingSF} is used, which is the same as the original paper.
  \item WSeq \cite{Tian2017HowTM}: The research in \cite{Tian2017HowTM} verifies the capability of the hierarchical models and 
  proposes a hierarchical model WSeq, which explicitly weights context vectors by context-query relevance (cosine similarity).
  \item HRAN \cite{Xing2017HierarchicalRA}: HRAN contains a hierarchical recurrent attention mechanism to fully leverage the context information, called word-level and utterance-level attention.
  \item DSHRED \cite{Zhang2018ContextSensitiveGO}: DSHRED uses dynamic and static attention mechanisms to generate more context-sensitive responses. 
  The query of the dynamic attention is the decoder hidden state, and the query of the static attention is the hidden state of the last utterance in the conversation.
  \item ReCoSa \cite{Zhang2019ReCoSaDT}: ReCoSa leverages the multi-head self-attention to detect multiple relative utterances in the context and achieves the state-of-the-art performance. 
  We run the original codes of the ReCoSa\footnote{\url{https://github.com/zhanghainan/ReCoSa}}, but it fails to generate any meaningful responses and is very hard to converge. 
  In this paper, we replace the transformer decoder of ReCoSa with the RNN-based decoder and only use the multi-head self-attention in the encoder, which is more stable and effective than original version ReCoSa.
\end{itemize}

Noted that there are also some hierarchical models based on HRED model, like Dir-VHRED \cite{Zeng2019DirichletLV}.
Because the improvements of these works are small, and they are not representative.
Thus, we don't choose them in this paper, which will not affect the conclusions.

\noindent \textbf{Non-hierarchical Models}:  
The non-hierarchical models usually make use of the Seq2Seq architecture, 
which is mainly divided into the following two kinds:
\begin{itemize}
  \item Seq2Seq+attn \cite{Cho2014LearningPR}: Seq2Seq architecture is first proposed for neural machine translation, 
  which is also widely for constructing the generative open-domain dialog systems. 
  Our implementation is consistent with the original paper, and the GRUs are used as the component of encoder and decoder modules.
  Besides, the attention module is also added \cite{Bahdanau2014NeuralMT}.
  \item Seq2Seq+trs \cite{Vaswani2017AttentionIA}: The transformer model leverages the multi-head self-attention mechanism to model the very long conversation context, 
  and shows very powerful performance on various natural language generation tasks. 
  In our experiments, we find that the vanilla transformer model is very sensitive to the hyper-parameters and the its performance is very unstable, which fails to achieve good performance.
  So we modify the vanilla transformer model by applying the multi-head self-attention module on the Seq2Seq+attn model.
\end{itemize}

Noted that some transformer-based non-hierarchical models are proposed recently, like DialoGPT \cite{Zhang2019DialoGPTLG}. 
They are essentially the same as the Seq2Seq+trs model. Thus, in this paper, we only evaluate the Seq2Seq+trs model.

\subsubsection{Chosen Metrics}
In this paper, we choose the following automatic evaluations to measure the performance of the hierarchical and non-hierarchical models.
It should be noted that we don't choose the word-overlap-based metrics such as BLEU and ROUGE, because some researches \cite{Liu2016HowNT,Zhang2018ContextSensitiveGO} show that these metrics are unsatisfied for evaluating the open-domain dialog generation.
Recently, researches tend to leverage the human evaluation to evaluate the open-domain multi-turn dialog systems. 
However, the human evaluations are very expensive, time-consuming and irreproducible.
Besides, there are lots of comparisons for the hierarchical and non-hierarchical models, so it is unpractical for us to collect the human annotations in this work.
Thus, in addition to the most commonly used automatic evaluations, we also introduce two state-of-the-art automatic evaluations to accurately measure the performance of these models, called BERT-RUBER and BERTScore. 

\begin{itemize} 
  \item Dist-1/2 \cite{Li2015ADO}: The two metrics dist-1/2 measure the degree of the diversity of the responses by calculating the number of distinct unigrams and bigrams. 
  If the two metrics are low, the responses generated by the models are very likely to be the safe responses, i.e., the bad, meaningless responses.
  \item Embedding-based metrics \cite{Liu2016HowNT}: Embedding-based metrics measure the performance by calculating the similarity of sentence embeddings between the ground-truth and the generated response. 
  In this paper, we use \textbf{Embedding Average}, \textbf{Vector Extrema} and \textbf{Greedy Matching} methods.
  Hereafter, we call them \textbf{Average, Extrema, Greedy}.
  Besides, the GoogleNews word2vec is used as the word embeddings\footnote{\url{https://github.com/mmihaltz/word2vec-GoogleNews-vectors}}.
  \item BERTScore \cite{Zhang2019BERTScoreET}: BERTScore leverages the tf-idf weighted BERT embedding to calculate the semantic similarity between responses and ground-truths,
  which shows the very powerful capability to measure the quality of the natural language.
  In this work, the latest BERTScore version 0.3.0 is used in this work\footnote{\url{https://github.com/Tiiiger/bert\_score}}.
  \item BERT-RUBER \cite{Tao2017RUBERAU}: Extensive experiments \cite{Liu2016HowNT} already demonstrate that the word-overlap-based metrics (BLEU, ROUGE) and embedding-based metrics show the relatively low correlation with human judgments.
  So a learning-based metric \cite{Tao2017RUBERAU} is proposed, which is trained by negative sampling, and shows a very high correlation with human judgments, called RUBER.
  Furthermore, BERT-RUBER \cite{Ghazarian2019BetterAE} leverages the BERT contextual word embedding to improve the performance of RUBER and shows the closest correlation with human judgments.
  In this work, we implement the BERT-RUBER and already released the codes\footnote{The anonymous link: \url{https://github.com/anonymous/xxx}}.
\end{itemize}

\subsubsection{Parameter settings}
All of the models are implemented by PyTorch.
For all the models, the early stopping mechanism and the dropout \cite{Srivastava2014DropoutAS} are used to avoid overfitting.
It should be noted that GRUs \cite{Cho2014LearningPR} are used for all the RNN cells.
All hyperparameter settings can be found in Table \ref{tab:3}, 
which is consistent among all of the chosen hierarchical models and non-hierarchical models.

\begin{table}[H]
  \begin{center}
    \resizebox{0.4\textwidth}{!}{
    \begin{tabular}{cc}
      \toprule[2pt]
      \multicolumn{1}{c}{\textbf{Hyperparameter}} & \textbf{Value} \\ \hline
      Learning rate schedule                        &  ReduceLROnPlateau \\
      Learning rate decay ratio                     &  0.5           \\
      Patience of LR decay                          &  10            \\
      GRU hidden size                               &  512           \\
      Utterance encoder layer  & 2 \\
      Bidirectional encoder  & True \\
      Context encoder layer & 1 \\
      Decoder layer & 2 \\
      Gradient clip & 3.0 \\
      Learning rate & 1e-4 \\
      Optimizer & Adam \\
      Dropout ratio & 0.3 \\
      Weight decay & 1e-6 \\
      Epochs & 100 \\
      Word embed size & 256 \\
      Seed & 30 \\ \hline
      Multi-head & 8 \\
      $d_{model}$ & 512 \\
      Transformer layer & 3 \\
      \bottomrule[2pt]
    \end{tabular}
    }
    \caption{Hyperparameter setting.}
    \label{tab:3}
  \end{center}
\end{table}

\subsection{Hierarchical vs. Non-hierarchical}

\begin{table*}[t]
  \begin{center}
    \subtable[Results on DailyDialog dataset.]{
    \begin{tabular}{|c|c|c|c|c|c|c|c|}
    \toprule[2pt]
    \hline
    \textbf{Model}        & \textbf{Dist-1} & \textbf{Dist-2} & \textbf{Average} & \textbf{Extrema} & \textbf{Greedy} & \textbf{BERTScore} & \textbf{BERT-RUBER} \\ \hline
    \textbf{Seq2Seq+attn} & 2.25            & 10.77           & 61.63            & 77.06            & 49.80           & 13.75              & 57.09          \\ \hline
    \textbf{Seq2Seq+trs}  & 2.64 $\ddag$    & 13.88$\dag$     & 62.53 $\ddag$    & 77.45 $\ddag$    & 51.29 $\ddag$   & 15.76$\dag$        & 64.23 $\ddag$  \\ \hline \hline
    \textbf{HRED}         & 1.46            & 6.71            & 60.27            & 76.09            & 49.08           & 13.41              & 58.81          \\ \hline
    \textbf{WSeq}         & 1.17            & 5.46            & 60.97            & 76.41            & 48.92           & 13.10              & 59.60          \\ \hline
    \textbf{VHRED}        & 1.62            & 6.90            & 59.90            & 75.50            & 48.65           & 12.27              & 60.62          \\ \hline
    \textbf{DSHRED}       & 1.85            & 9.05            & 61.39            & 76.87            & 49.67           & 14.43              & 62.10          \\ \hline
    \textbf{ReCoSa}       & 2.19            & 10.82           & 62.51            & 77.54$\dag$      & 50.81           & 15.67 $\ddag$      & 62.53          \\ \hline \hline
    \textbf{HRAN}         & 2.67$\dag$      & 13.71 $\ddag$   & 62.88$\dag$      & 77.36            & 51.33$\dag$     & 15.61              & 66.42$\dag$    \\ \hline
    \bottomrule[2pt]
    \end{tabular}
    }
  \end{center}
  \qquad
  \begin{center}
    \subtable[Results on PersonaChat dataset.]{
    \begin{tabular}{|c|c|c|c|c|c|c|c|}
    \toprule[2pt]
    \hline
    \textbf{Model}        & \textbf{Dist-1} & \textbf{Dist-2} & \textbf{Average} & \textbf{Extrema} & \textbf{Greedy} & \textbf{BERTScore} & \textbf{BERT-RUBER} \\ \hline
    \textbf{Seq2Seq+attn} & 0.79 $\dag$     & 4.57 $\ddag$    & 63.36            & 83.52            & 48.87           & 16.41 $\ddag$      & 41.05          \\ \hline
    \textbf{Seq2Seq+trs}  & 0.77 $\ddag$    & 4.76 $\dag$     & 64.65 $\dag$     & 83.74 $\dag$     & 49.38 $\ddag$   & 16.03              & 42.48 $\ddag$  \\ \hline \hline
    \textbf{HRED}         & 0.34            & 1.98            & 62.76            & 83.30            & 48.46           & 15.82              & 40.74          \\ \hline
    \textbf{WSeq}         & 0.41            & 2.41            & 63.25            & 83.53 $\ddag$    & 48.43           & 15.96              & 42.06          \\ \hline
    \textbf{VHRED}        & 0.50            & 2.78            & 63.04            & 83.40            & 48.61           & 16.29              & 41.59          \\ \hline
    \textbf{DSHRED}       & 0.44            & 2.63            & 63.02            & 83.44            & 48.73           & 15.96              & 42.16          \\ \hline
    \textbf{ReCoSa}       & 0.47            & 2.93            & 63.29            & 83.31            & 48.56           & 15.65              & 41.84          \\ \hline \hline
    \textbf{HRAN}         & 0.67            & 4.04            & 63.65 $\ddag$    & 83.47            & 49.50 $\dag$    & 16.99 $\dag$       & 42.97 $\dag$   \\ \hline
    \bottomrule[2pt]
    \end{tabular}
    }
  \end{center}

    \caption{Automatic evaluation (\%) on DailyDialog and PersonaChat dataset. 
    $\dag$ represents the best performance, and $\ddag$ represents the second best performance.}
    \label{tab:4}
\end{table*}

\begin{figure}[t]
  \centering
  \subfigure[Results on DailyDialog dataset.]{
      \begin{minipage}[t]{\linewidth}
          \centering
          \includegraphics[width=7cm, height=5cm]{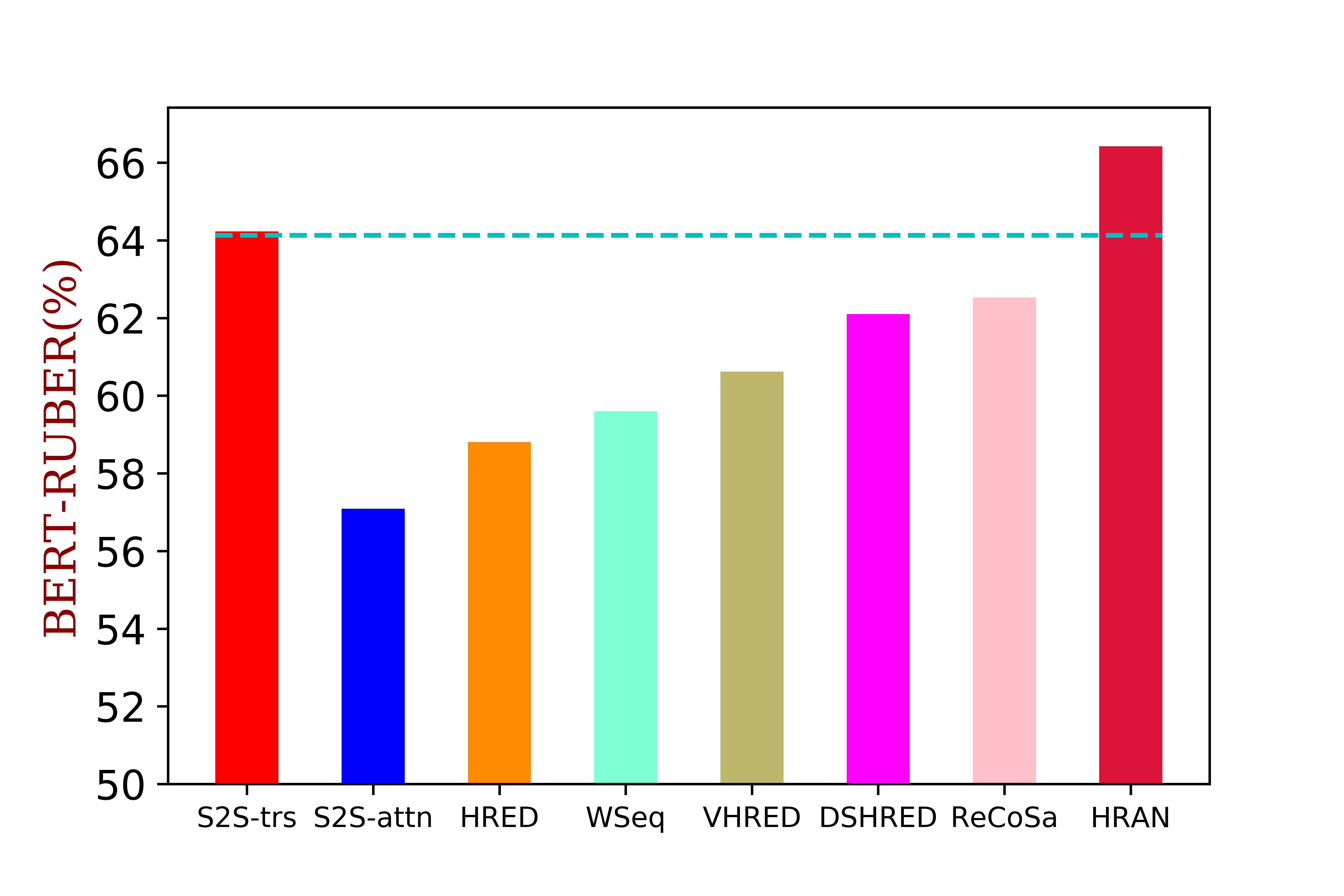}
      \end{minipage}%
      \label{img:2a}
  }%
  
  \subfigure[Results on PersonaChat dataset.]{
      \begin{minipage}[t]{\linewidth}
          \centering
          \includegraphics[width=7cm, height=5cm]{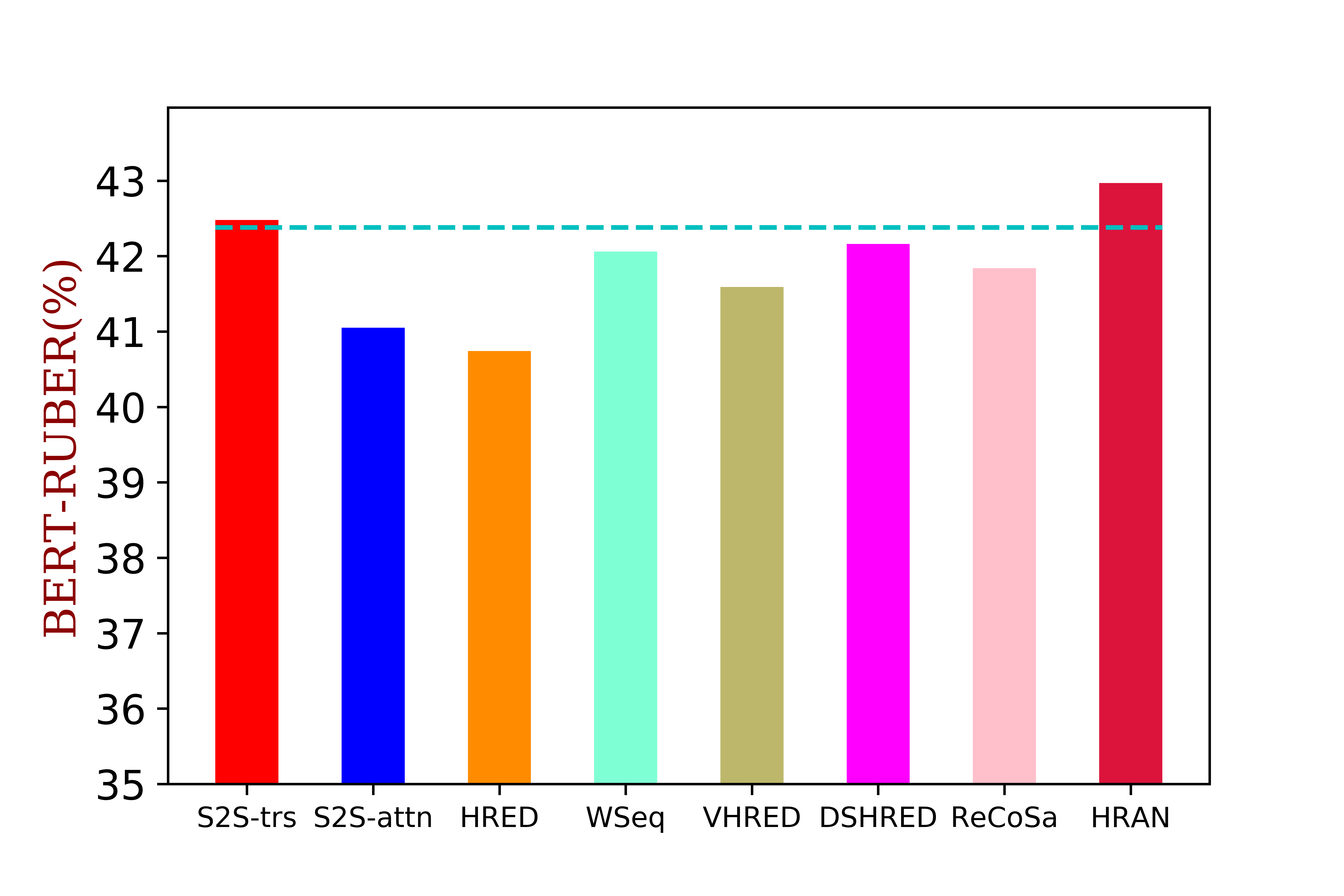}
      \end{minipage}%
      \label{img:2b}
  }%
  \centering
  \caption{Hierarchical models vs. Modified hierarchical models. 
  The best Seq2Seq results are reported.}
  \label{img:2}
\end{figure}

As shown in Figure \ref{img:2} and Table \ref{tab:4}, we can make the following conclusions:
\begin{itemize}
  \item As shown in Figure \ref{img:2}, it can be found that nearly all of the hierarchical models are worse than the non-hierarchical models, except for HRAN model.
  The worse performance of the hierarchical models is caused by the complicated hierarchical architecture, which makes them easily forget the essential fine-grained information, such as the valuable tokens.
  It should be noted that, ReCoSa and DSHRED perform better than non-hierarchical models in their papers, but they perform the worse performance in our experiments. 
  Maybe the reasons are as follows: 
  (1) The implementations of models in their works are different from our models, such as the different parameters and training settings. 
  In our work, the training and parameter settings of all the models are consistent to guarantee the fairness of the comparisons; 
  (2) In their experiments, they only use one or two datasets to evaluate the models, which may be insufficient;
  \item As shown in Figure \ref{img:2} and Table \ref{tab:4}, 
  it can be shown that the existing hierarchical models WSeq, VHRED, DSHRED, ReCoSa and HRAN are better than the basic hierarchical model HRED,
  which means that the modifications of HRED model are effective.
  \item As shown in Table \ref{tab:4}, it can be found that the HRAN model significantly outperforms other hierarchical models and non-hierarchical models on the state-of-the-art automatic metrics BERTScore and BERT-RUBER.
  The main difference between the HRAN model and other hierarchical models is that it has the word-level attention mechanism, which potentially illustrates its powerful capability.
\end{itemize}

\subsection{Does Word-level Attention Work Well for Other Hierarchical Models?}
Through the systematic comparisons of the hierarchical and non-hierarchical models, 
it can be found that word-level attention has the potential capability to improve the performance of the hierarchical models.
In order to check whether the word-level attention mechanism has the consistent improvement for the hierarchical architecture,
we add the word-level attention mechanism into other hierarchical models and show the comparisons in this section.

\begin{figure}[H]
  \centering
  \subfigure[Results on DailyDialog dataset.]{
      \begin{minipage}[t]{\linewidth}
          \centering
          \includegraphics[width=7cm, height=4.55cm]{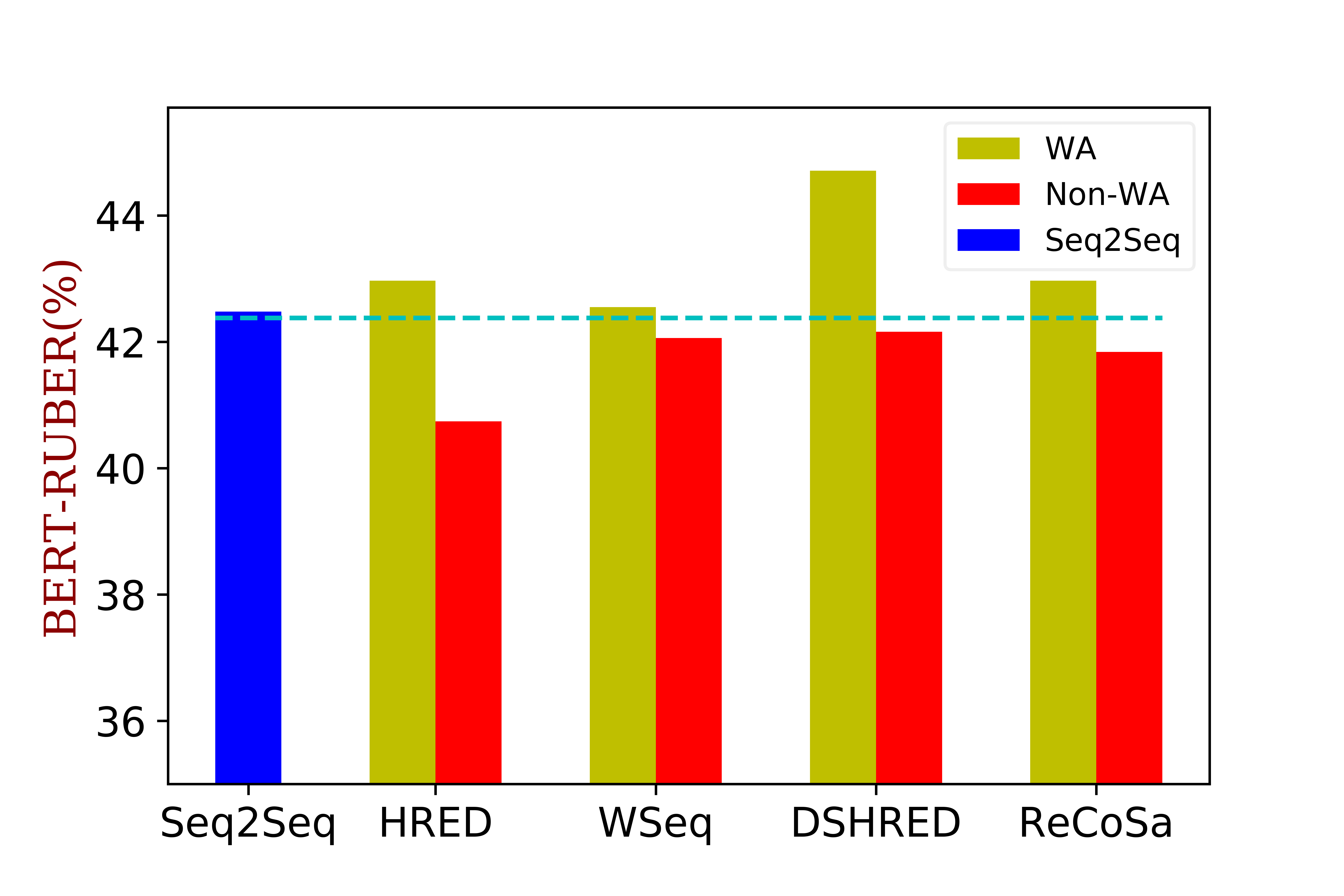}
      \end{minipage}%
      \label{img:3a}
  }%
  
  \subfigure[Results on EmpChat dataset.]{
      \begin{minipage}[t]{\linewidth}
          \centering
          \includegraphics[width=7cm, height=4.55cm]{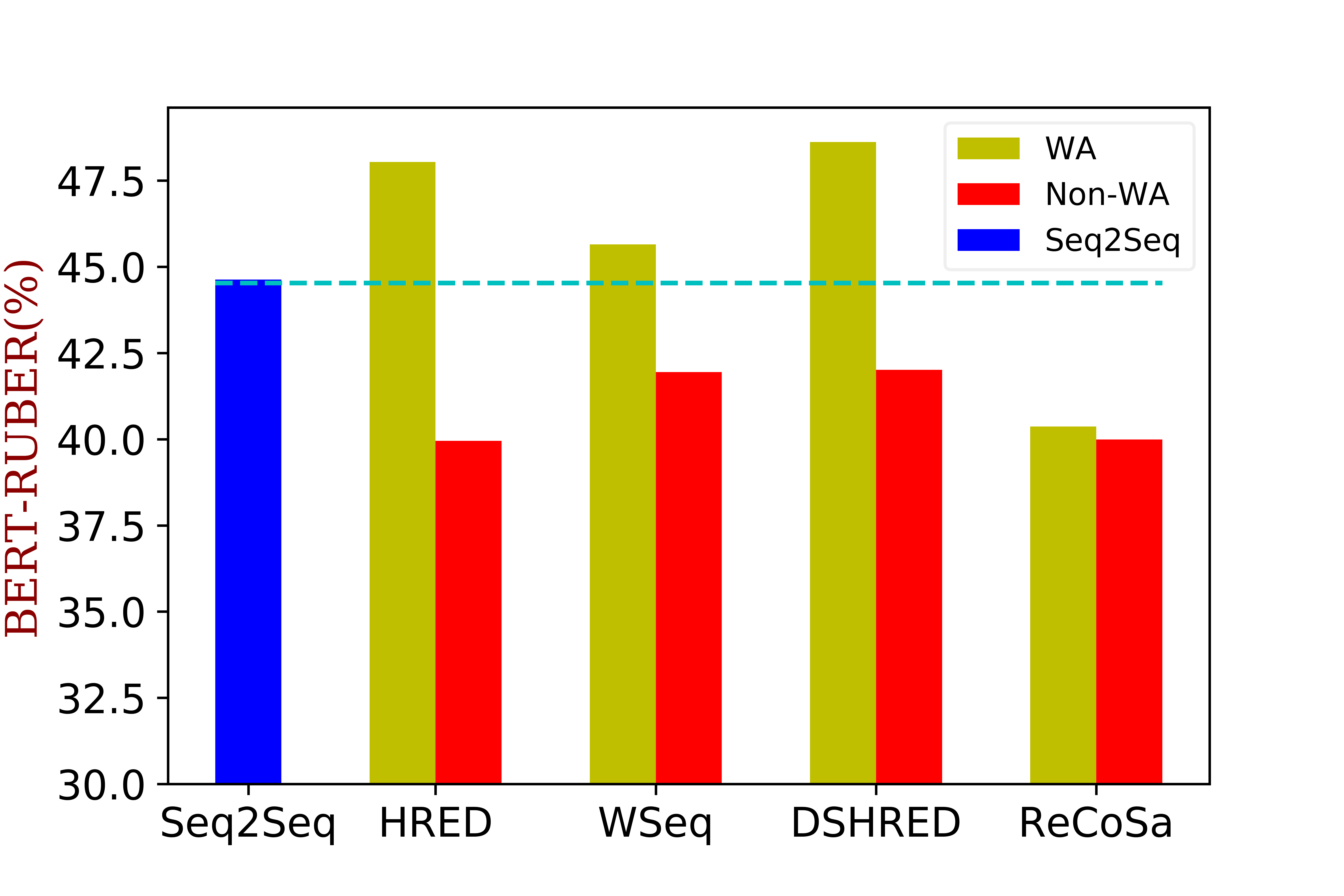}
      \end{minipage}%
      \label{img:3b}
  }%
  \centering
  \caption{Hierarchical models vs. Modified hierarchical models on DailyDialog and EmpChat dataset. 
  The best Seq2Seq results are shown in the blue bar.}
  \label{img:3}
\end{figure}

\subsubsection{Chosen Hierarchical models}
It should be noted that not all of the hierarchical models can add the word-level attention mechanism. 
For example, VHRED model cannot add word-level attention mechanism because of its latent variable mechanism. 
In this paper, we add the word-level attention mechanism into HRED, WSeq, DSHRED and ReCoSa models, and obtain the corresponding modified models.
For example, DSHRED+WA represents the DSHRED with word-level attention mechanism.

\subsubsection{Results}

\begin{table*}[t]
  \begin{center}
    \subtable[Results on DailyDialog dataset.]{
    \begin{tabular}{|c|c|c|c|c|c|c|c|}
    \toprule[2pt]
    \hline
    \textbf{Model}        & \textbf{Dist-1} & \textbf{Dist-2} & \textbf{Average} & \textbf{Extrema} & \textbf{Greedy} & \textbf{BERTScore} & \textbf{BERT-RUBER} \\ \hline
    \textbf{Seq2Seq+attn} & 2.25            & 10.77           & 61.63            & 77.06            & 49.80           & 13.75              & 57.09          \\ \hline
    \textbf{Seq2Seq+trs}  & 2.64            & 13.88           & 62.53 $\ddag$    & 77.45  $\dag$    & 51.29 $\ddag$   & 15.76 $\ddag$      & 64.23          \\ \hline \hline
    \textbf{HRED+WA}      & 2.67            & 13.71           & 62.88 $\dag$     & 77.36  $\ddag$   & 51.33           & 15.61              & 66.42          \\ \hline
    \textbf{WSeq+WA}      & 2.68            & 13.65           & 61.66            & 77.12            & 50.31           & 14.93              & 66.52 $\ddag$         \\ \hline
    \textbf{DSHRED+WA}    & 2.85  $\ddag$   & 14.66 $\ddag$   & 62.32            & 77.09            & 51.75 $\dag$    & 16.48 $\dag$       & 67.27 $\dag$         \\ \hline
    \textbf{ReCoSa+WA}    & 3.13  $\dag$    & 15.06 $\dag$    & 61.33            & 76.56            & 49.84           & 14.45              & 59.73          \\ \hline
    \bottomrule[2pt]
    \end{tabular}
    }
  \end{center}  

  \begin{center}
    \subtable[Results on PersonaChat dataset.]{
      \begin{tabular}{|c|c|c|c|c|c|c|c|}
      \toprule[2pt]
      \hline
      \textbf{Model}        & \textbf{Dist-1} & \textbf{Dist-2} & \textbf{Average} & \textbf{Extrema} & \textbf{Greedy} & \textbf{BERTScore} & \textbf{BERT-RUBER} \\ \hline
      \textbf{Seq2Seq+attn} & 0.79 $\ddag$    & 4.57            & 63.36            & 83.52            & 48.87           & 16.41              & 41.05               \\ \hline
      \textbf{Seq2Seq+trs}  & 0.77            & 4.76 $\ddag$    & 64.65 $\dag$     & 83.74 $\dag$     & 49.38           & 16.03              & 42.48               \\ \hline \hline
      \textbf{HRED+WA}      & 0.67            & 4.04            & 63.65            & 82.47            & 49.50 $\dag$    & 16.99  $\dag$      & 42.97 $\ddag$              \\ \hline
      \textbf{WSeq+WA}      & 0.59            & 3.20            & 62.94            & 83.43            & 48.69           & 16.52              & 42.55               \\ \hline
      \textbf{DSHRED+WA}    & 0.80 $\dag$     & 4.84 $\dag$     & 64.14 $\ddag$    & 83.64 $\ddag$    & 49.48 $\ddag$   & 16.68 $\ddag$      & 44.71 $\dag$       \\ \hline
      \textbf{ReCoSa+WA}    & 0.62            & 3.39            & 63.39            & 83.74 $\dag$     & 48.78           & 15.34              & 42.97 $\ddag$              \\ \hline
      \bottomrule[2pt]
      \end{tabular}
    }
  \end{center}
  \caption{Automatic evaluation on Dailydialog and PersonaChat dataset.
  $\dag$ represents the best performance, and $\ddag$ represents the second best performance.
  It can be found that modified hierarchical models are better than non-hierarchical models on most of the automatic evaluations.}
  \label{tab:5}
\end{table*}

The comparisons between the modified hierarchical models on BERT-RUBER metric are shown in Figure \ref{img:3},
and we can make the following conclusions:
\begin{itemize}
  \item Compared with the original hierarchical models (red bar),
  the word-level attention mechanism significantly improves the performance of modified models (yellow bar) on BERT-RUBER metric,
  which means that the word-level attention mechanism is effective for the hierarchical architecture.
  \item 
  Nearly all of the modified hierarchical models (yellow bar) significantly outperform the state-of-the-art non-hierarchical models (blue bar).
  \item The improvement of DSHRED model is the best significant.
  For example, 6.61\% and 5.17\% improvement can be achieved on EmpChat and Dailydialog datasets.
  \item Although the word-level attention mechanism indeed improves the ReCoSa model, 
  its improvement is not as large as the one of other hierarchical models.
  Through careful analysis, the reasons may be the incompatibility of the multi-head self-attention mechanism and vanilla attention mechanism.  
  In the future, we will study how to combine these two attention mechanisms more effectively.
\end{itemize}

Moreover, all of the automatic evaluations of these modified hierarchical models are shown in Table \ref{tab:5}.
Through these results, it can be observed:
(1) For most of the automatic evaluations, especially on BERTScore and BERT-RUBER, 
the word-level attention can greatly improve the hierarchical models,
which significantly outperform the non-hierarchical models;
(2) Compared with other modified hierarchical models, the DSHRED+WA model achieves the best performance on most of the automatic evaluations, especially the state-of-the-art metric BERT-RUBER.
Other modified hierarchical models don't outperform non-hierarchical models significantly.

\subsection{Why Word-level Attention Is So Effective? Quantitative Analysis}
Through the above discussion and analysis, 
it is clear that the word-level attention mechanism is very necessary for hierarchical models,
which can significantly improve their performance.
But it is still a confusing question why the word-level attention is so effective.

Intuitively, word-level attention can generate better utterance representations for the context encoder of the hierarchical models.
Besides, lots of fine-grained information, for example, the word-level information can be leveraged effectively.
In other words, the word-level attention can help the model leverage the information of the multi-turn conversation context effectively.
In the paper of HRAN \cite{Xing2017HierarchicalRA}, they only conduct the ablation study to analyze the contribution of the word-level attention, 
which is very coarse and insufficient.
But, how to quantitatively measure the model's capability of using contextual information?

The perturbation test \cite{Khandelwal2018SharpNF,Sankar2019DoND} is a novel and effective method to 
measure the capability of the generative models to utilize context information in natural language process.
The central premise of the perturbation test is that \textit{models make minimal use of certain
types of information if they are insensitive to perturbations that destroy them}.
Specifically, 10 kinds of perturbation are injected into the multi-turn conversation context only during the test stage \cite{Sankar2019DoND},
and the decrease of the performance is reported. 
If the performance of the models decrease badly, it means that the models effectively use the context information; otherwise not.
However, the original perturbation test \cite{Sankar2019DoND} uses the perplexity metric for evaluating the generative model's performance, which is unsuitable for dialogue modeling \cite{Liu2016HowNT}.
So in our work, we replace the perplexity with the state-of-the-art metrics BERTScore and BERT-RUBER, and the perturbations are as follows:

\begin{table*}[t]
  \begin{center}
  \begin{minipage}{\textwidth}
    \begin{minipage}[t]{0.5\textwidth}
        \centering  
        \makeatletter\def\makeatother\caption{BERT-RUBER}
        \resizebox{\textwidth}{!}{
        \begin{tabular}{|c|c|c|c|c|}
          \toprule[2pt]
          \hline
          \textbf{Model}       & \textbf{DailyDialog} & \textbf{EmpChat} & \textbf{PersonaChat} & \textbf{DSTC7-AVSD} \\ \hline \hline
          \textbf{Seq2Seq+trs} & \textbf{-7.46}       & -12.02           & -14.01               & -17.34              \\ \hline \hline
          \textbf{HRED}        & -5.83                & -11.55           & -13.80               & -17.29              \\ \hline
          \textbf{HRED+WA}     & -6.64 $\uparrow$     & \textbf{-14.10} $\uparrow$ & -17.15 $\uparrow$    & \textbf{-21.04} $\uparrow$   \\ \hline \hline
          \textbf{WSeq}        & -6.04                & -11.80           & -13.38               & -17.96              \\ \hline
          \textbf{WSeq+WA}     & -7.05 $\uparrow$     & -12.77 $\uparrow$& \textbf{-17.72} $\uparrow$    & -20.48 $\uparrow$   \\ \hline \hline
          \textbf{DSHRED}      & -7.38                & -11.81           & -15.00               & -17.23              \\ \hline
          \textbf{DSHRED+WA}   & -7.28 $\downarrow$   & -13.34 $\uparrow$& -14.93 $\downarrow$  & -17.29 $\uparrow$   \\ \hline \hline
          \textbf{ReCoSa}      & -6.67                & -11.74           & -14.49               & -17.08              \\ \hline
          \textbf{ReCoSa+WA}   & -5.98 $\downarrow$   & -12.20 $\uparrow$& -15.62 $\uparrow$    & -17.22 $\uparrow$   \\ \hline
          \bottomrule[2pt]
        \end{tabular}
        }
    \end{minipage}
    \begin{minipage}[t]{0.5\textwidth}
        \centering
        \makeatletter\def\makeatother\caption{BERTScore}
        \resizebox{\textwidth}{!}{
        \begin{tabular}{|c|c|c|c|c|}
          \toprule[2pt]
          \hline
          \textbf{Model}       & \textbf{DailyDialog} & \textbf{EmpChat} & \textbf{PersonaChat} & \textbf{DSTC7-AVSD} \\ \hline \hline
          \textbf{Seq2Seq+trs} & -5.37                & -2.01            & -2.96                & -8.29               \\ \hline \hline
          \textbf{HRED}        & -3.55                & -1.69            & -4.08                & -9.04               \\ \hline
          \textbf{HRED+WA}     & -4.93 $\uparrow$     & -2.10 $\uparrow$ & -4.13 $\uparrow$     & -9.13  $\uparrow$   \\ \hline \hline
          \textbf{WSeq}        & -3.15                & -1.91            & -2.92                & -6.73               \\ \hline
          \textbf{WSeq+WA}     & -4.33  $\uparrow$    & \textbf{-2.14} $\uparrow$ & -4.71 $\uparrow$     & \textbf{-9.68} $\uparrow$    \\ \hline \hline
          \textbf{DSHRED}      & -4.00                & -2.02            & -3.54                & -7.62               \\ \hline
          \textbf{DSHRED+WA}   & \textbf{-5.84}  $\uparrow$    & -2.11 $\uparrow$ & \textbf{-4.88} $\uparrow$     & -9.63 $\uparrow$    \\ \hline \hline
          \textbf{ReCoSa}      & -4.97                & -1.42            & -3.18                & -7.33               \\ \hline
          \textbf{ReCoSa+WA}   & -4.50  $\downarrow$  & -1.42 $\sim$     & -2.50 $\downarrow$   & -6.96 $\downarrow$  \\ \hline
          \bottomrule[2pt]
        \end{tabular}
        }
    \end{minipage}
  \end{minipage}
  \caption{Perturbation test on four datasets. It should be noted that the scores (\%) are the average decrease in performance.
  The higher the decrease of the performance, the more effective the model is in leveraging the multi-turn context. 
  $\uparrow$ means the capability of using the context information is better, $\downarrow$ means the worse capability. 
  The best results are shown in bold. The left table shows the results on the BERT-RUBER metric, and the right table shows the results on BERTScore metric.}
  \label{tab:6}
  \end{center}
\end{table*}

\subsubsection{Utterance-level perturbation}
\begin{itemize}
  \item \textit{Shuffle}: shuffles the sequence of utterances in the multi-turn dialog history.
  \item \textit{Reverse}: Reverses the order of utterances in the history (but maintain the word order within each utterance).
  \item \textit{Drop first}: Drops the first sentence in the dialog history.
  \item \textit{Drop last}: Drops the last sentence e.g. query in the dialog history.
  \item \textit{Truncate}: truncates the dialog history to contain only the $k$ most recent utterance where $k \leq n$, 
  where $n$ is the number of the utterances in the multi-turn conversation. In this paper, the $k$ is 1 e.g. 
  only contain the last utterance as the conversation context.
\end{itemize}

\subsubsection{Word-level perturbation}
\begin{itemize}
  \item \textit{Word shuffle}: randomly shuffles the words within each utterance.
  \item \textit{Word reverse}: reverses the ordering of the words in each utterance.
  \item \textit{Word drop}: drops 30\% of the words uniformly in each utterance.
  \item \textit{Noun drop}: drops all the nouns.
  \item \textit{Verb drop}: drops all the verbs.
\end{itemize}

The performance decrease on BERT-RUBER and BERTScore caused by the perturbation are shown in Table \ref{tab:6},
and the average decrease of the performance on 10 perturbations are reported.
From these results, we can make the following conclusions:
(1) it can be found that the performance decrease of most modified models are \textbf{better than original hierarchical models},
which means that they leverage the multi-turn context information more effectively than original hierarchical models;
(2) In most cases, the modified hierarchical models are more sensitive to the perturbations \textbf{than non-hierarchical models},
which means that modified hierarchical models leverage the context information more effectively than non-hierarchical models. 
This also explains why the modified hierarchical models significantly outperform the non-hierarchical models.

The quantitative analysis from perturbation test demonstrates that modified hierarchical models leverage the multi-turn context more effectively,
which explains why the word-level attention mechanism is necessary for hierarchical models.

\begin{table}[h]
  \begin{center}
  \resizebox{0.48\textwidth}{!}{
    \begin{tabular}{|c|c|}
    \toprule[2pt]
    \hline
      \textbf{Context}      &  \makecell[c]{A man is standing against the wall of a house or building \\ when \textbf{another} man walks up from the side \\ The man who walks up is carrying a bag and can be heard \\ There is a man that is holding a cellphone and texting \\ A man then enters and places things in a counter \\ How many persons are in the video ?} \\ \hline \hline
      \textbf{Ground-Truth}  &  \makecell[c]{There are \textbf{two} people in the video} \\ \hline
      \textbf{Seq2Seq+trs}   &  \makecell[c]{There is \textbf{only one} person in the video .} \\ \hline
      \textbf{HRED}      &  \makecell[c]{There is \textbf{one} man in the video .} \\ \hline
      \textbf{HRED+WA}  &  \makecell[c]{There are \textbf{two} people in the video .} \\ \hline
      \bottomrule[2pt]
    \end{tabular}
  }
  \caption{Real examples in the DSTC7-AVSD datasets. The generated responses of Seq2Seq+trs, HRED and HRED+WA models are reported.
  In this example, the keyword \textit{another} is very valuable information for the question.}
  \label{tab:7}
  \end{center}
\end{table}

\begin{figure*}[t]
  \centering
  \subfigure[Heatmap of HRED+WA model.]{
      \begin{minipage}[t]{\linewidth}
          \centering
          \includegraphics[width=18cm, height=3.4cm]{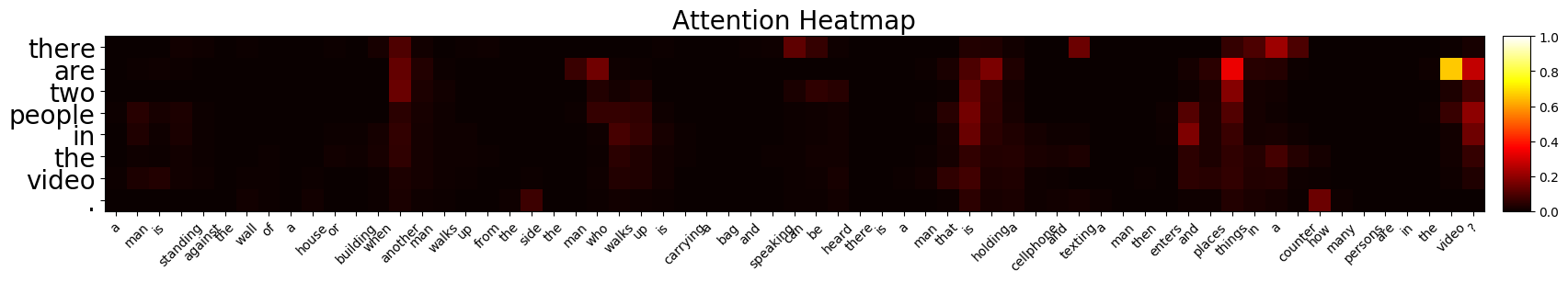}
      \end{minipage}%
      \label{img:5a}
  }%
  
  \subfigure[Heatmap of Non-hierarchical Seq2Seq+trs.]{
      \begin{minipage}[t]{\linewidth}
          \centering
          \includegraphics[width=18cm, height=3.4cm]{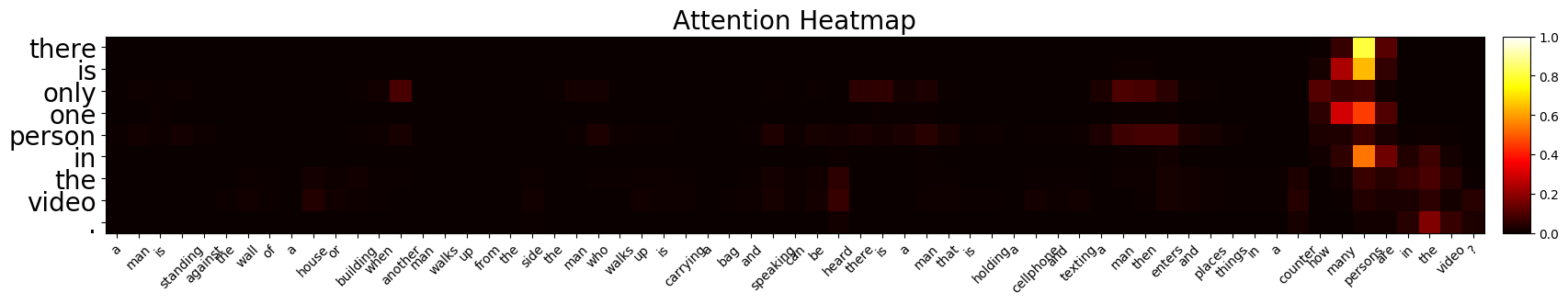}
      \end{minipage}%
      \label{img:5b}
  }%
  \centering
  \caption{Attention hearmap of the real examples on DSTC7-AVSD dataset.}
  \label{img:5}
\end{figure*}

\subsection{Why Word-level Attention Is So Effective? Qualitative Analysis}
In this section, in order to qualitatively show the capability of the word-level attention mechanism,
we show the attention weights heatmap of the HRED+WA, HRED and Seq2Seq+trs models.
We use a real example from the DSTC7-AVSD dataset to show the attention scores, and the example is shown in Table \ref{tab:7}.
In this example, the query is to ask the dialog model a question \textit{How many persons are in the video}. 
The model needs to make full use of the information in the context and give the correct answer \textit{two people}.
Besides, the most valuable utterance is the first sentence which contains the keyword \textit{another}.
As shown in Table \ref{tab:7}, it can be found that only HRED+WA model provides the correct answer,
which means that other models use context information incorrectly or inadequately.

\begin{figure}[t]
  \centering
  \subfigure[Heatmap of HRED+WA.]{
      \begin{minipage}[t]{0.5\linewidth}
          \centering
          \includegraphics[width=4cm, height=5cm]{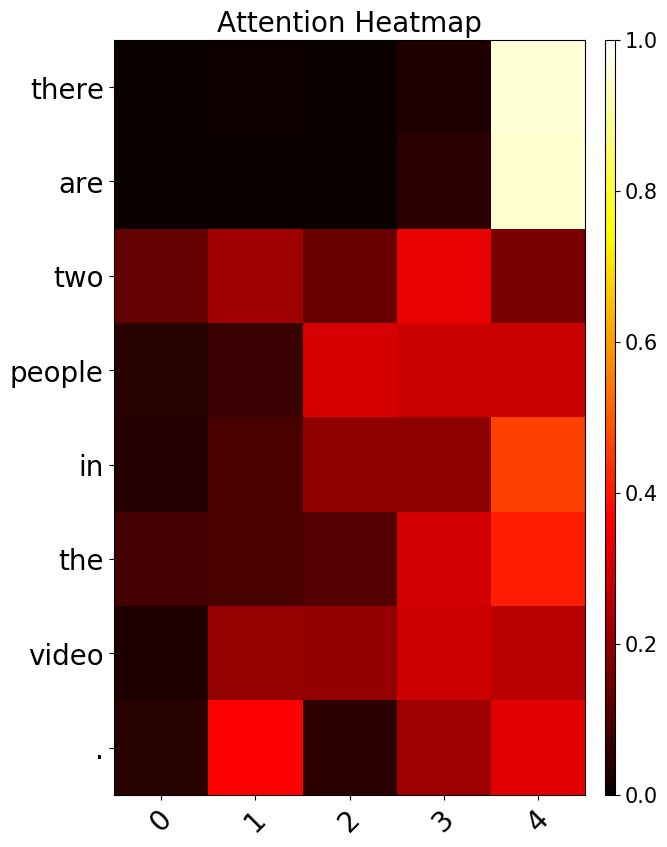}
      \end{minipage}%
      \label{img:4a}
  }%
  \subfigure[Heatmap of HRED.]{
      \begin{minipage}[t]{0.5\linewidth}
          \centering
          \includegraphics[width=4cm, height=5cm]{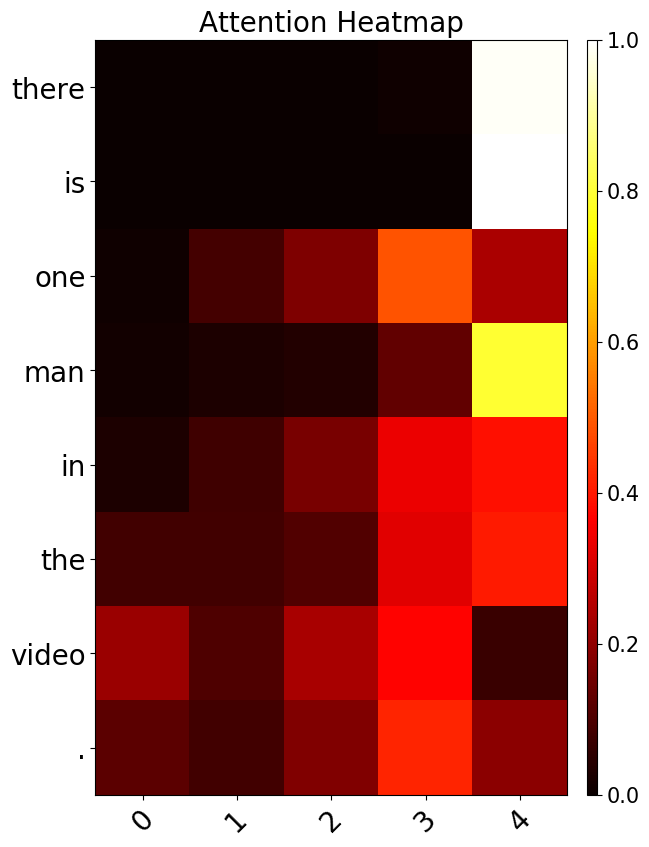}
      \end{minipage}%
      \label{img:4b}
  }%
  \centering
  \caption{Context-level attention scores on DSTC7 dataset. 
  The numbers in X-axis represent multiple sentences in the context.}
  \label{img:4}
\end{figure}

Firstly, we conduct comparisons between the HRED and HRED+  WA models.
In these comparisons, the third token in the response is very important, for example, the token \textit{one} leads to the wrong result.
In order to qualitatively analyze why the HRED decides to generate the token \textit{one}, the context-level attention scores are shown in Figure \ref{img:4} (b).
It can be found that HRED model ignores the information of the first sentence, and the attention score of the first utterance is 8.9e-3.
The reason of HRED's unsatisfied performance may be as follows:
(1) The complicated hierarchical architecture makes the HRED easily forgets the essential word-level information;
(2) The utterance representations generated by word-level encoder are unsatisfied.
By contrast, the HRED+WA can generate more appropriate attention scores. 
The attention score of the first utterance is 0.134, which is much higher than the one in HRED.
Because the HRED+WA model collects the essential word-level information and generate better utterance representations,
HRED+WA can focus on valuable utterance and generate the correct answer.

Then, we also conduct the comparisons between the HRED+WA and Seq2Seq+trs models.
The attention heatmaps are shown in Figure \ref{img:5}.
It can be found that the non-hierarchical model Seq2Seq+trs tends to leverage the nearby context, and barely focuses on the valuable first utterance far away. 
This phenomenon is already reported by the recent works \cite{Sankar2019DoND,Khandelwal2018SharpNF}.
The HRED+WA model, in contrast, can effectively focus on the fine-grained word-level information \textit{another} in the first utterance.
Specifically, the attention score of HRED+WA is 0.0504, which is much higher than 7.4e-4 of Seq2Seq+trs model.
Compared with the non-hierarchical model, the modified hierarchical models make full use of the context information, 
and focus on the fine-grained information, to generate more context-sensitive responses.

\section{Conclusions and future work} 
Open-domain multi-turn generative dialog generation is a big challenge.
Due to the lack of adequate and systematic comparisons between these two kinds of models,
it is still not clear which kind of models are better in open-domain multi-turn dialog generation.
Thus, in this paper, we measure systematically nearly all representative hierarchical and non-hierarchical models over the same experimental setting to check which kind is better.

Extensive experiments demonstrate three important conclusions:
(1) Nearly all the hierarchical models are worse than non-hierarchical models, except for the HRAN model, which contains the word-level attention mechanism;
(2) The performance of hierarchical models can be greatly improved by integrating word-level attention mechanism.
Besides, the modified hierarchical models even significantly outperform the state-of-the-art non-hierarchical models;
(3) The reason why the word-level attention mechanism is so powerful for hierarchical models is because it can leverage context more effectively, especially on fine-grained information.

Although extensive experiments demonstrate that the word-level attention mechanism is very important for hierarchical models, it still has some fatal weaknesses.
For example, training and inferernce stage of the modified hierarchical models is slower than the hierarchical models without word-level attention mechanism.
During decoding every token, the context-level encoder needs to re-process the utterance representations generated by the word-level attention, which is very time-consuming.
So in the future, we would like to improve the efficiency of the word-level attention mechanism and accelerate the training and inference stage.

\bibliographystyle{ACM-Reference-Format}
\bibliography{sample-base}


\begin{thebibliography}{32}


\ifx \showCODEN    \undefined \def \showCODEN     #1{\unskip}     \fi
\ifx \showDOI      \undefined \def \showDOI       #1{#1}\fi
\ifx \showISBNx    \undefined \def \showISBNx     #1{\unskip}     \fi
\ifx \showISBNxiii \undefined \def \showISBNxiii  #1{\unskip}     \fi
\ifx \showISSN     \undefined \def \showISSN      #1{\unskip}     \fi
\ifx \showLCCN     \undefined \def \showLCCN      #1{\unskip}     \fi
\ifx \shownote     \undefined \def \shownote      #1{#1}          \fi
\ifx \showarticletitle \undefined \def \showarticletitle #1{#1}   \fi
\ifx \showURL      \undefined \def \showURL       {\relax}        \fi
\providecommand\bibfield[2]{#2}
\providecommand\bibinfo[2]{#2}
\providecommand\natexlab[1]{#1}
\providecommand\showeprint[2][]{arXiv:#2}

\bibitem[\protect\citeauthoryear{Adiwardana, Luong, So, Hall, Fiedel,
  Thoppilan, Yang, Kulshreshtha, Nemade, Lu, and Le}{Adiwardana
  et~al\mbox{.}}{2020}]%
        {Adiwardana2020TowardsAH}
\bibfield{author}{\bibinfo{person}{Daniel De~Freitas Adiwardana},
  \bibinfo{person}{Minh-Thang Luong}, \bibinfo{person}{David~R. So},
  \bibinfo{person}{Jamie Hall}, \bibinfo{person}{Noah Fiedel},
  \bibinfo{person}{Romal Thoppilan}, \bibinfo{person}{Zi Yang},
  \bibinfo{person}{Apoorv Kulshreshtha}, \bibinfo{person}{Gaurav Nemade},
  \bibinfo{person}{Yifeng Lu}, {and} \bibinfo{person}{Quoc~V. Le}.}
  \bibinfo{year}{2020}\natexlab{}.
\newblock \showarticletitle{Towards a Human-like Open-Domain Chatbot}.
\newblock \bibinfo{journal}{\emph{ArXiv}}  \bibinfo{volume}{abs/2001.09977}
  (\bibinfo{year}{2020}).
\newblock


\bibitem[\protect\citeauthoryear{AlAmri, Cartillier, Das, Wang, Lee, Anderson,
  Essa, Parikh, Batra, Cherian, Marks, and Hori}{AlAmri et~al\mbox{.}}{2019}]%
        {AlAmri2019AudioVS}
\bibfield{author}{\bibinfo{person}{Huda AlAmri}, \bibinfo{person}{Vincent
  Cartillier}, \bibinfo{person}{Abhishek Das}, \bibinfo{person}{Jue Wang},
  \bibinfo{person}{Stefan Lee}, \bibinfo{person}{Pip Anderson},
  \bibinfo{person}{Irfan Essa}, \bibinfo{person}{Devi Parikh},
  \bibinfo{person}{Dhruv Batra}, \bibinfo{person}{Anoop Cherian},
  \bibinfo{person}{Tim~K. Marks}, {and} \bibinfo{person}{Chiori Hori}.}
  \bibinfo{year}{2019}\natexlab{}.
\newblock \showarticletitle{Audio Visual Scene-Aware Dialog}.
\newblock \bibinfo{journal}{\emph{2019 IEEE/CVF Conference on Computer Vision
  and Pattern Recognition (CVPR)}} (\bibinfo{year}{2019}),
  \bibinfo{pages}{7550--7559}.
\newblock


\bibitem[\protect\citeauthoryear{Bahdanau, Cho, and Bengio}{Bahdanau
  et~al\mbox{.}}{2014}]%
        {Bahdanau2014NeuralMT}
\bibfield{author}{\bibinfo{person}{Dzmitry Bahdanau},
  \bibinfo{person}{Kyunghyun Cho}, {and} \bibinfo{person}{Yoshua Bengio}.}
  \bibinfo{year}{2014}\natexlab{}.
\newblock \showarticletitle{Neural Machine Translation by Jointly Learning to
  Align and Translate}.
\newblock \bibinfo{journal}{\emph{CoRR}}  \bibinfo{volume}{abs/1409.0473}
  (\bibinfo{year}{2014}).
\newblock


\bibitem[\protect\citeauthoryear{Bowman, Vilnis, Vinyals, Dai, J{\'o}zefowicz,
  and Bengio}{Bowman et~al\mbox{.}}{2015}]%
        {Bowman2015GeneratingSF}
\bibfield{author}{\bibinfo{person}{Samuel~R. Bowman}, \bibinfo{person}{Luke
  Vilnis}, \bibinfo{person}{Oriol Vinyals}, \bibinfo{person}{Andrew~M. Dai},
  \bibinfo{person}{Rafal J{\'o}zefowicz}, {and} \bibinfo{person}{Samy Bengio}.}
  \bibinfo{year}{2015}\natexlab{}.
\newblock \showarticletitle{Generating Sentences from a Continuous Space}. In
  \bibinfo{booktitle}{\emph{CoNLL}}.
\newblock


\bibitem[\protect\citeauthoryear{Chen, Ren, Tang, Zhao, and Yin}{Chen
  et~al\mbox{.}}{2018}]%
        {Chen2018HierarchicalVM}
\bibfield{author}{\bibinfo{person}{Hongshen Chen}, \bibinfo{person}{Zhaochun
  Ren}, \bibinfo{person}{Jiliang Tang}, \bibinfo{person}{Yihong~Eric Zhao},
  {and} \bibinfo{person}{Dawei Yin}.} \bibinfo{year}{2018}\natexlab{}.
\newblock \showarticletitle{Hierarchical Variational Memory Network for
  Dialogue Generation}. In \bibinfo{booktitle}{\emph{WWW}}.
\newblock


\bibitem[\protect\citeauthoryear{Cho, van Merrienboer, Çaglar G{\"u}lçehre,
  Bahdanau, Bougares, Schwenk, and Bengio}{Cho et~al\mbox{.}}{2014}]%
        {Cho2014LearningPR}
\bibfield{author}{\bibinfo{person}{Kyunghyun Cho}, \bibinfo{person}{Bart van
  Merrienboer}, \bibinfo{person}{Çaglar G{\"u}lçehre},
  \bibinfo{person}{Dzmitry Bahdanau}, \bibinfo{person}{Fethi Bougares},
  \bibinfo{person}{Holger Schwenk}, {and} \bibinfo{person}{Yoshua Bengio}.}
  \bibinfo{year}{2014}\natexlab{}.
\newblock \showarticletitle{Learning Phrase Representations using RNN
  Encoder-Decoder for Statistical Machine Translation}.
\newblock \bibinfo{journal}{\emph{ArXiv}}  \bibinfo{volume}{abs/1406.1078}
  (\bibinfo{year}{2014}).
\newblock


\bibitem[\protect\citeauthoryear{Ghazarian, Wei, Galstyan, and Peng}{Ghazarian
  et~al\mbox{.}}{2019}]%
        {Ghazarian2019BetterAE}
\bibfield{author}{\bibinfo{person}{Sarik Ghazarian}, \bibinfo{person}{Johnny
  Tian-Zheng Wei}, \bibinfo{person}{Aram Galstyan}, {and}
  \bibinfo{person}{Nanyun Peng}.} \bibinfo{year}{2019}\natexlab{}.
\newblock \showarticletitle{Better Automatic Evaluation of Open-Domain Dialogue
  Systems with Contextualized Embeddings}.
\newblock \bibinfo{journal}{\emph{ArXiv}}  \bibinfo{volume}{abs/1904.10635}
  (\bibinfo{year}{2019}).
\newblock


\bibitem[\protect\citeauthoryear{Hochreiter and Schmidhuber}{Hochreiter and
  Schmidhuber}{1997}]%
        {Hochreiter1997LongSM}
\bibfield{author}{\bibinfo{person}{Sepp Hochreiter} {and}
  \bibinfo{person}{J{\"u}rgen Schmidhuber}.} \bibinfo{year}{1997}\natexlab{}.
\newblock \showarticletitle{Long Short-Term Memory}.
\newblock \bibinfo{journal}{\emph{Neural Computation}}  \bibinfo{volume}{9}
  (\bibinfo{year}{1997}), \bibinfo{pages}{1735--1780}.
\newblock


\bibitem[\protect\citeauthoryear{Khandelwal, He, Qi, and Jurafsky}{Khandelwal
  et~al\mbox{.}}{2018}]%
        {Khandelwal2018SharpNF}
\bibfield{author}{\bibinfo{person}{Urvashi Khandelwal}, \bibinfo{person}{He
  He}, \bibinfo{person}{Peng Qi}, {and} \bibinfo{person}{Dan Jurafsky}.}
  \bibinfo{year}{2018}\natexlab{}.
\newblock \showarticletitle{Sharp Nearby, Fuzzy Far Away: How Neural Language
  Models Use Context}. In \bibinfo{booktitle}{\emph{ACL}}.
\newblock


\bibitem[\protect\citeauthoryear{Li, Galley, Brockett, Gao, and Dolan}{Li
  et~al\mbox{.}}{2015}]%
        {Li2015ADO}
\bibfield{author}{\bibinfo{person}{Jiwei Li}, \bibinfo{person}{Michel Galley},
  \bibinfo{person}{Chris Brockett}, \bibinfo{person}{Jianfeng Gao}, {and}
  \bibinfo{person}{William~B. Dolan}.} \bibinfo{year}{2015}\natexlab{}.
\newblock \showarticletitle{A Diversity-Promoting Objective Function for Neural
  Conversation Models}.
\newblock \bibinfo{journal}{\emph{ArXiv}}  \bibinfo{volume}{abs/1510.03055}
  (\bibinfo{year}{2015}).
\newblock


\bibitem[\protect\citeauthoryear{Li, Su, Shen, Li, Cao, and Niu}{Li
  et~al\mbox{.}}{2017}]%
        {Li2017DailyDialogAM}
\bibfield{author}{\bibinfo{person}{Yanran Li}, \bibinfo{person}{Hui Su},
  \bibinfo{person}{Xiaoyu Shen}, \bibinfo{person}{Wenjie Li},
  \bibinfo{person}{Ziqiang Cao}, {and} \bibinfo{person}{Shuzi Niu}.}
  \bibinfo{year}{2017}\natexlab{}.
\newblock \showarticletitle{DailyDialog: A Manually Labelled Multi-turn
  Dialogue Dataset}. In \bibinfo{booktitle}{\emph{IJCNLP}}.
\newblock


\bibitem[\protect\citeauthoryear{Liu, Lowe, Serban, Noseworthy, Charlin, and
  Pineau}{Liu et~al\mbox{.}}{2016}]%
        {Liu2016HowNT}
\bibfield{author}{\bibinfo{person}{Chia-Wei Liu}, \bibinfo{person}{Ryan Lowe},
  \bibinfo{person}{Iulian Serban}, \bibinfo{person}{Michael Noseworthy},
  \bibinfo{person}{Laurent Charlin}, {and} \bibinfo{person}{Joelle Pineau}.}
  \bibinfo{year}{2016}\natexlab{}.
\newblock \showarticletitle{How NOT To Evaluate Your Dialogue System: An
  Empirical Study of Unsupervised Evaluation Metrics for Dialogue Response
  Generation}.
\newblock \bibinfo{journal}{\emph{ArXiv}}  \bibinfo{volume}{abs/1603.08023}
  (\bibinfo{year}{2016}).
\newblock


\bibitem[\protect\citeauthoryear{Rashkin, Smith, Li, and Boureau}{Rashkin
  et~al\mbox{.}}{2019}]%
        {rashkin2019towards}
\bibfield{author}{\bibinfo{person}{Hannah Rashkin},
  \bibinfo{person}{Eric~Michael Smith}, \bibinfo{person}{Margaret Li}, {and}
  \bibinfo{person}{Y-Lan Boureau}.} \bibinfo{year}{2019}\natexlab{}.
\newblock \showarticletitle{Towards Empathetic Open-domain Conversation Models:
  a New Benchmark and Dataset}. In \bibinfo{booktitle}{\emph{ACL}}.
\newblock


\bibitem[\protect\citeauthoryear{Sankar, Subramanian, Pal, Chandar, and
  Bengio}{Sankar et~al\mbox{.}}{2019}]%
        {Sankar2019DoND}
\bibfield{author}{\bibinfo{person}{Chinnadhurai Sankar}, \bibinfo{person}{S.
  Subramanian}, \bibinfo{person}{Christopher~Joseph Pal},
  \bibinfo{person}{A.~P.~Sarath Chandar}, {and} \bibinfo{person}{Yoshua
  Bengio}.} \bibinfo{year}{2019}\natexlab{}.
\newblock \showarticletitle{Do Neural Dialog Systems Use the Conversation
  History Effectively? An Empirical Study}. In \bibinfo{booktitle}{\emph{ACL}}.
\newblock


\bibitem[\protect\citeauthoryear{Serban, Sordoni, Bengio, Courville, and
  Pineau}{Serban et~al\mbox{.}}{2015}]%
        {Serban2015BuildingED}
\bibfield{author}{\bibinfo{person}{Iulian Serban}, \bibinfo{person}{Alessandro
  Sordoni}, \bibinfo{person}{Yoshua Bengio}, \bibinfo{person}{Aaron~C.
  Courville}, {and} \bibinfo{person}{Joelle Pineau}.}
  \bibinfo{year}{2015}\natexlab{}.
\newblock \showarticletitle{Building End-To-End Dialogue Systems Using
  Generative Hierarchical Neural Network Models}. In
  \bibinfo{booktitle}{\emph{AAAI}}.
\newblock


\bibitem[\protect\citeauthoryear{Serban, Sordoni, Lowe, Charlin, Pineau,
  Courville, and Bengio}{Serban et~al\mbox{.}}{2016}]%
        {Serban2016AHL}
\bibfield{author}{\bibinfo{person}{Iulian Serban}, \bibinfo{person}{Alessandro
  Sordoni}, \bibinfo{person}{Ryan Lowe}, \bibinfo{person}{Laurent Charlin},
  \bibinfo{person}{Joelle Pineau}, \bibinfo{person}{Aaron~C. Courville}, {and}
  \bibinfo{person}{Yoshua Bengio}.} \bibinfo{year}{2016}\natexlab{}.
\newblock \showarticletitle{A Hierarchical Latent Variable Encoder-Decoder
  Model for Generating Dialogues}. In \bibinfo{booktitle}{\emph{AAAI}}.
\newblock


\bibitem[\protect\citeauthoryear{Shang, Lu, and Li}{Shang
  et~al\mbox{.}}{2015}]%
        {Shang2015NeuralRM}
\bibfield{author}{\bibinfo{person}{Lifeng Shang}, \bibinfo{person}{Zhengdong
  Lu}, {and} \bibinfo{person}{Hang Li}.} \bibinfo{year}{2015}\natexlab{}.
\newblock \showarticletitle{Neural Responding Machine for Short-Text
  Conversation}. In \bibinfo{booktitle}{\emph{ACL}}.
\newblock


\bibitem[\protect\citeauthoryear{Srivastava, Hinton, Krizhevsky, Sutskever, and
  Salakhutdinov}{Srivastava et~al\mbox{.}}{2014}]%
        {Srivastava2014DropoutAS}
\bibfield{author}{\bibinfo{person}{Nitish Srivastava},
  \bibinfo{person}{Geoffrey~E. Hinton}, \bibinfo{person}{Alex Krizhevsky},
  \bibinfo{person}{Ilya Sutskever}, {and} \bibinfo{person}{Ruslan
  Salakhutdinov}.} \bibinfo{year}{2014}\natexlab{}.
\newblock \showarticletitle{Dropout: a simple way to prevent neural networks
  from overfitting}.
\newblock \bibinfo{journal}{\emph{J. Mach. Learn. Res.}}  \bibinfo{volume}{15}
  (\bibinfo{year}{2014}), \bibinfo{pages}{1929--1958}.
\newblock


\bibitem[\protect\citeauthoryear{Sutskever, Vinyals, and Le}{Sutskever
  et~al\mbox{.}}{2014}]%
        {Sutskever2014SequenceTS}
\bibfield{author}{\bibinfo{person}{Ilya Sutskever}, \bibinfo{person}{Oriol
  Vinyals}, {and} \bibinfo{person}{Quoc~V. Le}.}
  \bibinfo{year}{2014}\natexlab{}.
\newblock \showarticletitle{Sequence to Sequence Learning with Neural
  Networks}. In \bibinfo{booktitle}{\emph{NIPS}}.
\newblock


\bibitem[\protect\citeauthoryear{Tao, Mou, Zhao, and Yan}{Tao
  et~al\mbox{.}}{2017}]%
        {Tao2017RUBERAU}
\bibfield{author}{\bibinfo{person}{Chongyang Tao}, \bibinfo{person}{Lili Mou},
  \bibinfo{person}{Dongyan Zhao}, {and} \bibinfo{person}{Rui Yan}.}
  \bibinfo{year}{2017}\natexlab{}.
\newblock \showarticletitle{RUBER: An Unsupervised Method for Automatic
  Evaluation of Open-Domain Dialog Systems}.
\newblock \bibinfo{journal}{\emph{ArXiv}}  \bibinfo{volume}{abs/1701.03079}
  (\bibinfo{year}{2017}).
\newblock


\bibitem[\protect\citeauthoryear{Tian, Yan, Mou, Song, Feng, and Zhao}{Tian
  et~al\mbox{.}}{2017}]%
        {Tian2017HowTM}
\bibfield{author}{\bibinfo{person}{Zhiliang Tian}, \bibinfo{person}{Rui Yan},
  \bibinfo{person}{Lili Mou}, \bibinfo{person}{Yiping Song},
  \bibinfo{person}{Yansong Feng}, {and} \bibinfo{person}{Dongyan Zhao}.}
  \bibinfo{year}{2017}\natexlab{}.
\newblock \showarticletitle{How to Make Context More Useful? An Empirical Study
  on Context-Aware Neural Conversational Models}. In
  \bibinfo{booktitle}{\emph{ACL}}.
\newblock


\bibitem[\protect\citeauthoryear{Vaswani, Shazeer, Parmar, Uszkoreit, Jones,
  Gomez, Kaiser, and Polosukhin}{Vaswani et~al\mbox{.}}{2017}]%
        {Vaswani2017AttentionIA}
\bibfield{author}{\bibinfo{person}{Ashish Vaswani}, \bibinfo{person}{Noam
  Shazeer}, \bibinfo{person}{Niki Parmar}, \bibinfo{person}{Jakob Uszkoreit},
  \bibinfo{person}{Llion Jones}, \bibinfo{person}{Aidan~N. Gomez},
  \bibinfo{person}{Lukasz Kaiser}, {and} \bibinfo{person}{Illia Polosukhin}.}
  \bibinfo{year}{2017}\natexlab{}.
\newblock \showarticletitle{Attention is All you Need}. In
  \bibinfo{booktitle}{\emph{NIPS}}.
\newblock


\bibitem[\protect\citeauthoryear{Wang, Huang, Xu, Shen, and Nie}{Wang
  et~al\mbox{.}}{2018}]%
        {Wang2018ChatMD}
\bibfield{author}{\bibinfo{person}{Wenjie Wang}, \bibinfo{person}{Minlie
  Huang}, \bibinfo{person}{Xin-Shun Xu}, \bibinfo{person}{Fumin Shen}, {and}
  \bibinfo{person}{Liqiang Nie}.} \bibinfo{year}{2018}\natexlab{}.
\newblock \showarticletitle{Chat More: Deepening and Widening the Chatting
  Topic via A Deep Model}. In \bibinfo{booktitle}{\emph{SIGIR '18}}.
\newblock


\bibitem[\protect\citeauthoryear{Xing, Wu, Wu, Zhou, Huang, and Ma}{Xing
  et~al\mbox{.}}{2017}]%
        {Xing2017HierarchicalRA}
\bibfield{author}{\bibinfo{person}{Chen Xing}, \bibinfo{person}{Wei~Yu Wu},
  \bibinfo{person}{Yu Wu}, \bibinfo{person}{Ming Zhou}, \bibinfo{person}{Yalou
  Huang}, {and} \bibinfo{person}{Wei-Ying Ma}.}
  \bibinfo{year}{2017}\natexlab{}.
\newblock \showarticletitle{Hierarchical Recurrent Attention Network for
  Response Generation}.
\newblock \bibinfo{journal}{\emph{ArXiv}}  \bibinfo{volume}{abs/1701.07149}
  (\bibinfo{year}{2017}).
\newblock


\bibitem[\protect\citeauthoryear{Xu, Sun, Long, Liu, Wang, Wang, Zhang, and
  Wang}{Xu et~al\mbox{.}}{2019}]%
        {Xu2019DynamicWM}
\bibfield{author}{\bibinfo{person}{Zhen Xu}, \bibinfo{person}{Chengjie Sun},
  \bibinfo{person}{Yinong Long}, \bibinfo{person}{Bingquan Liu},
  \bibinfo{person}{Baoxun Wang}, \bibinfo{person}{Mingjiang Wang},
  \bibinfo{person}{Min Zhang}, {and} \bibinfo{person}{Xiaolong Wang}.}
  \bibinfo{year}{2019}\natexlab{}.
\newblock \showarticletitle{Dynamic Working Memory for Context-Aware Response
  Generation}.
\newblock \bibinfo{journal}{\emph{IEEE/ACM Transactions on Audio, Speech, and
  Language Processing}}  \bibinfo{volume}{27} (\bibinfo{year}{2019}),
  \bibinfo{pages}{1419--1431}.
\newblock


\bibitem[\protect\citeauthoryear{Zeng, Wang, and Luo}{Zeng
  et~al\mbox{.}}{2019}]%
        {Zeng2019DirichletLV}
\bibfield{author}{\bibinfo{person}{Min Zeng}, \bibinfo{person}{Yisen Wang},
  {and} \bibinfo{person}{Yuan Luo}.} \bibinfo{year}{2019}\natexlab{}.
\newblock \showarticletitle{Dirichlet Latent Variable Hierarchical Recurrent
  Encoder-Decoder in Dialogue Generation}. In
  \bibinfo{booktitle}{\emph{EMNLP/IJCNLP}}.
\newblock


\bibitem[\protect\citeauthoryear{Zhang, Lan, Pang, Guo, and Cheng}{Zhang
  et~al\mbox{.}}{2019b}]%
        {Zhang2019ReCoSaDT}
\bibfield{author}{\bibinfo{person}{Hemmg Zhang}, \bibinfo{person}{Yanyan Lan},
  \bibinfo{person}{Liang Pang}, \bibinfo{person}{Jiafeng Guo}, {and}
  \bibinfo{person}{Xueqi Cheng}.} \bibinfo{year}{2019}\natexlab{b}.
\newblock \showarticletitle{ReCoSa: Detecting the Relevant Contexts with
  Self-Attention for Multi-turn Dialogue Generation}. In
  \bibinfo{booktitle}{\emph{ACL}}.
\newblock


\bibitem[\protect\citeauthoryear{Zhang, Dinan, Urbanek, Szlam, Kiela, and
  Weston}{Zhang et~al\mbox{.}}{2018b}]%
        {Zhang2018PersonalizingDA}
\bibfield{author}{\bibinfo{person}{Saizheng Zhang}, \bibinfo{person}{Emily
  Dinan}, \bibinfo{person}{Jack Urbanek}, \bibinfo{person}{Arthur Szlam},
  \bibinfo{person}{Douwe Kiela}, {and} \bibinfo{person}{Jason Weston}.}
  \bibinfo{year}{2018}\natexlab{b}.
\newblock \showarticletitle{Personalizing Dialogue Agents: I have a dog, do you
  have pets too?}. In \bibinfo{booktitle}{\emph{ACL}}.
\newblock


\bibitem[\protect\citeauthoryear{Zhang, Kishore, Wu, Weinberger, and
  Artzi}{Zhang et~al\mbox{.}}{2019a}]%
        {Zhang2019BERTScoreET}
\bibfield{author}{\bibinfo{person}{Tianyi Zhang}, \bibinfo{person}{Varsha
  Kishore}, \bibinfo{person}{Felix Wu}, \bibinfo{person}{Kilian~Q. Weinberger},
  {and} \bibinfo{person}{Yoav Artzi}.} \bibinfo{year}{2019}\natexlab{a}.
\newblock \showarticletitle{BERTScore: Evaluating Text Generation with BERT}.
\newblock \bibinfo{journal}{\emph{ArXiv}}  \bibinfo{volume}{abs/1904.09675}
  (\bibinfo{year}{2019}).
\newblock


\bibitem[\protect\citeauthoryear{Zhang, Cui, Wang, Zhu, Li, Zhou, and
  Liu}{Zhang et~al\mbox{.}}{2018a}]%
        {Zhang2018ContextSensitiveGO}
\bibfield{author}{\bibinfo{person}{Weinan Zhang}, \bibinfo{person}{Yiming Cui},
  \bibinfo{person}{Yifa Wang}, \bibinfo{person}{Qingfu Zhu},
  \bibinfo{person}{Lingzhi Li}, \bibinfo{person}{Lianqiang Zhou}, {and}
  \bibinfo{person}{Ting Liu}.} \bibinfo{year}{2018}\natexlab{a}.
\newblock \showarticletitle{Context-Sensitive Generation of Open-Domain
  Conversational Responses}. In \bibinfo{booktitle}{\emph{COLING}}.
\newblock


\bibitem[\protect\citeauthoryear{Zhang, Sun, Galley, Chen, Brockett, Gao, Gao,
  Liu, and Dolan}{Zhang et~al\mbox{.}}{2019c}]%
        {Zhang2019DialoGPTLG}
\bibfield{author}{\bibinfo{person}{Yizhe Zhang}, \bibinfo{person}{Siqi Sun},
  \bibinfo{person}{Michel Galley}, \bibinfo{person}{Yen-Chun Chen},
  \bibinfo{person}{Chris Brockett}, \bibinfo{person}{Xiang Gao},
  \bibinfo{person}{Jianfeng Gao}, \bibinfo{person}{Jingjing Liu}, {and}
  \bibinfo{person}{William~B. Dolan}.} \bibinfo{year}{2019}\natexlab{c}.
\newblock \showarticletitle{DialoGPT: Large-Scale Generative Pre-training for
  Conversational Response Generation}.
\newblock \bibinfo{journal}{\emph{ArXiv}}  \bibinfo{volume}{abs/1911.00536}
  (\bibinfo{year}{2019}).
\newblock


\bibitem[\protect\citeauthoryear{Zhou, Young, Huang, Zhao, Xu, and Zhu}{Zhou
  et~al\mbox{.}}{2018}]%
        {Zhou2018CommonsenseKA}
\bibfield{author}{\bibinfo{person}{Hao Zhou}, \bibinfo{person}{Tom Young},
  \bibinfo{person}{Minlie Huang}, \bibinfo{person}{Haizhou Zhao},
  \bibinfo{person}{Jingfang Xu}, {and} \bibinfo{person}{Xiaoyan Zhu}.}
  \bibinfo{year}{2018}\natexlab{}.
\newblock \showarticletitle{Commonsense Knowledge Aware Conversation Generation
  with Graph Attention}. In \bibinfo{booktitle}{\emph{IJCAI}}.
\newblock


\end{thebibliography}


\end{document}